%% file: main.tex
\documentclass[final]{cvpr}

\usepackage{times}
\usepackage{epsfig}
\usepackage{graphicx}
\usepackage{amsmath}
\usepackage{amssymb}
\usepackage{kotex}
\usepackage[usenames,dvipsnames]{xcolor}
\usepackage{enumerate}  
\usepackage{enumitem}   
\usepackage{booktabs}
\usepackage{multirow}
\usepackage{pifont}
\usepackage{wrapfig}
\usepackage{subcaption}
\usepackage{float}
\usepackage[labelsep=period]{caption}

\usepackage[pagebackref=true,breaklinks=true,colorlinks,bookmarks=false]{hyperref}

\def\cvprPaperID{5043} %
\def\confYear{CVPR 2021}

\newcommand*{\mb}[1]{\mathbf{#1}}

\def\be#1\ee{\begin{align}#1\end{align}}
\def\bea#1\eea{\begin{eqnarray}#1\end{eqnarray}}
\def\ba#1\ea{\begin{align*}#1\end{align*}}
\def\bs#1\es{\begin{equation}\begin{split}#1\end{split}\end{equation}}
\newcommand{\stimes}{{\times}}

\hypersetup{colorlinks=true, linkcolor=blue, anchorcolor=blue, citecolor=blue, filecolor = blue, urlcolor = blue} 

\title{Rethinking Channel Dimensions for Efficient Model Design}%

\begin{document}

\author{Dongyoon Han \quad Sangdoo Yun \quad Byeongho Heo\quad YoungJoon Yoo\\ \\ 
 NAVER AI Lab
}

\maketitle
\input{00.Abstract.tex}

\input{01.Introduction.tex}

\input{02.Related_Work}
\input{03-1.Preliminary.tex}

\input{03-2.Problem_Formulation.tex}

\input{04.Network_Improvement.tex}

\input{05.Experiment.tex}
\input{06.Discussion.tex}

\input{07.Conclusion.tex}

\vspace{0.1cm}
{\small
\noindent\textbf{Acknowledgement}
We would like to thank NAVER AI Labs members including Sanghyuk Chun, Seong Joon Oh, and Junsuk Choe for fruitful discussions and peer-reviews. We also thank Jung-Woo Ha who suggested the name of the architecture and the NAVER Smart Machine Learning (NSML)~\cite{nsml} team for support.  %
}

{\small
\bibliographystyle{ieee_fullname}
\bibliography{egbib}
}

\clearpage
\section*{Appendix}

\input{appendix.tex}

\end{document}

%% file: 00.Abstract.tex
\begin{abstract}
Designing an efficient model within the limited computational cost is challenging. We argue the accuracy of a lightweight model has been further limited by the design convention: a stage-wise configuration of the channel dimensions, which looks like a piecewise linear function of the network stage. In this paper, we study an effective channel dimension configuration towards better performance than the convention. To this end, we empirically study how to design a single layer properly by analyzing the rank of the output feature. We then investigate the channel configuration of a model by searching network architectures concerning the channel configuration under the computational cost restriction. Based on the investigation, we propose a simple yet effective channel configuration that can be parameterized by the layer index. As a result, our proposed model following the channel parameterization achieves remarkable performance on ImageNet classification and transfer learning tasks including COCO object detection, COCO instance segmentation, and fine-grained classifications. Code and ImageNet pretrained models are available at {\url{https://github.com/clovaai/rexnet}}.
\end{abstract}
\vspace{-2mm}

%% file: 01.Introduction.tex
\input{tables/channel_config_table.tex}

\section{Introduction}
Designing a lightweight network architecture is crucial for both researchers and practitioners. 
Popular networks~\cite{VGG,resnet,mobilenetv1,mobilenetv2} designed for ImageNet classification share a similar design convention where a low dimensional input channel is expanded by a few channel expansion layers towards surpassing the number of classes. Lightweight models~\cite{mobilenetv1,mobilenetv2,mobilenetv3,shufflenetv1,shufflenetv2,mnasnet,fbnet, proxylessnas, efficientnet} also follow this configuration but further shrinks some channels for computational efficiency, which leads to the promising trade-offs between the computational cost and accuracy. 
In other words, the degree of channel expansion at layers is quite different, where earlier layers have a smaller channel dimension; the penultimate layer that largely expands dimension above the number of classes. 
This is to realize flop-efficiency by narrow channel dimensions at earlier layers; to get model expressiveness with sufficient channel dimension at the final feature (see Table~\ref{table:channel_dim_setting}).

This channel configuration was firstly introduced by MobileNetV2~\cite{mobilenetv2} and became the design convention of configuring channel dimensions in lightweight networks, but how to adjust the channel dimensions towards the optimal under the restricted computational cost has not been profoundly studied. As shown in Table~\ref{table:channel_dim_setting}, even network architecture search (NAS)-based models~\cite{fbnet, proxylessnas,mobilenetv3,mnasnet, tan2019mixconv,efficientnet,mei2020atomnas, chu2019fairnas, chu2019fairdarts, liang2019dartsplus} were designed upon the convention or little more exploration within few options near the configuration~\cite{chamnet, fbnetv2} and focused on searching building blocks. Take one step further from the design convention, we hypothesize that the compact models designed by the conventional channel configuration may be limited in the expressive power due to mainly focusing on flop-efficiency; there would exist a more effective configuration over the traditional one.

In this paper, we investigate an effective channel configuration of a lightweight network with additional accuracy gain. Inspired by the works~\cite{yang2017breaking, zhang2015efficient}, we conjecture the expressiveness of a layer can be estimated by the matrix rank of the output feature. Technically, we study with the averaged rank computed from the output feature of a bunch of networks that are randomly generated with random sizes to reveal the proper range of expansion ratio at an expansion layer and make a link with a rough design principle. %
Based on the principle, we move forward to find out an overall channel configuration in a network. Specifically, we search network architectures to identify the channel configuration yielding a better accuracy over the aforementioned convention. It turns out that the best channel configuration is parameterized as a linear function by the block index in a network. This parameterization is similar to the configuration used in the works~\cite{densenet, pyramidnet}, and we reveal the parameterization is also effective in designing a lightweight model.

Based on the investigation, we propose a new model upon the searched channel parameterization. It turns out that a simple modification upon MobileNetV2 could show remarkable improvement in performance on ImageNet classification. Only with the new channel configuration, our models outperform the state-of-the-art networks such as EfficientNets~\cite{efficientnet} whose architectures were found by the compound scaling with TPUs. 
This stresses the effectiveness of our channel configuration over the convention and may encourage the researchers in the NAS field to adopt our channel configuration into the network search space for further performance boosts. The performance improvement of ImageNet classification is well transferred to the object detection and instance segmentation on the COCO dataset~\cite{coco2017} and the various fine-grained classification tasks. This indicates our backbones work as strong feature extractors.

Our contributions are 1) a study on designing a single layer (\S\ref{section:sec3}); 2) a network architecture exploration concerning the channel configuration towards a simple yet effective parameterization (\S\ref{section:sec4}); 3) using our models to achieve remarkable results on ImageNet~\cite{imagenet} outperformed recent lightweight models including NAS-based models (\S\ref{section:exp}); 4) revealing the high applicability of our ImageNet-pretrained backbones transferring to several tasks including object detection, instance segmentation and fine-grained classification, which indicate the high expressiveness of our model and the effectiveness of our channel configuration (\S\ref{section:exp}).

%% file: tables/channel_config_table.tex
\begin{table*}[t]
\fontsize{8.1}{9.5}\selectfont
\centering
\tabcolsep=0.10cm
\hspace{-1mm}
\subfloat{\begin{tabular}{@{}l|l|c|c|c}
Network &  Stem / Building blocks' output channel dimensions &  Top-1 & Params & Flops\\
\midrule
MobileNetV2~\cite{mobilenetv2} & 32 / 16(×1)-24(×2)-32(×3)-64(×4)-96(×3)-160(×3)-320(×1) & 72.0\% & 3.4M & 0.30B  \\
FBNet-C~\cite{fbnet} & 16 / 16(×1)-24(×4)-32(×4)-64(×4)-112(×4)-184(×4)-352(×1) &  74.9\% & 5.5M & 0.38B  \\
ProxylessNas-R~\cite{proxylessnas} & 32 / 16(×1)-32(×2)-40(×4)-80(×4)-96(×4)-192(×4)-320(×1)  & 74.6\% & 4.1M & 0.32B  \\
MNasNet-A1~\cite{mnasnet} & 32 / 16(×1)-24(×2)-40(×3)-80(×4)-112(×2)-160(×3)-320(×1)  & 75.2\% & 3.9M & 0.31B  \\
MixNet-M~\cite{tan2019mixconv} & 24 / 24(×1)-32(×2)-40(×4)-80(×4)-120(×4)-200(×4) & 77.0\% & 5.0M & 0.36B  \\
EfficinetNet-B0~\cite{efficientnet} & 32 / 16(×1)-24(×2)-40(×2)-80(×3)-112(×3)-192(×4)-320(×1) & 77.3\% & 4.8M & 0.39B  \\
AtomNas-C~\cite{mei2020atomnas} & 32 / 16(×1)-24(×4)-40(×4)-80(×4)-96(×4)-192(×4)-320(×1)& 77.6\% & 5.9M & 0.36B  \\
FairNas-A~\cite{chu2019fairnas} & 32 / 16(×1)-32(×2)-40(×4)-80(×4)-96(×4)-192(×4)-320(×1)& 77.5\% &  5.9M & 0.39B \\
FairDARTS-C~\cite{chu2019fairdarts} & 32 / 16(×1)-32(×2)-40(×4)-80(×5)-96(×3)-192(×4)-320(×1)& 77.2\% & 5.3M & 0.39B  \\
SE-DARTS+~\cite{liang2019dartsplus} & 32 / 16(×1)-24(×4)-40(×4)-80(×4)-96(×4)-192(×4)-320(×1)& 77.5\% & 6.1M & 0.59B  \\ 
\end{tabular}}
\ \quad
\subfloat{
\includegraphics[trim = 1mm 2mm 0mm 0mm, clip, width=0.26\linewidth]{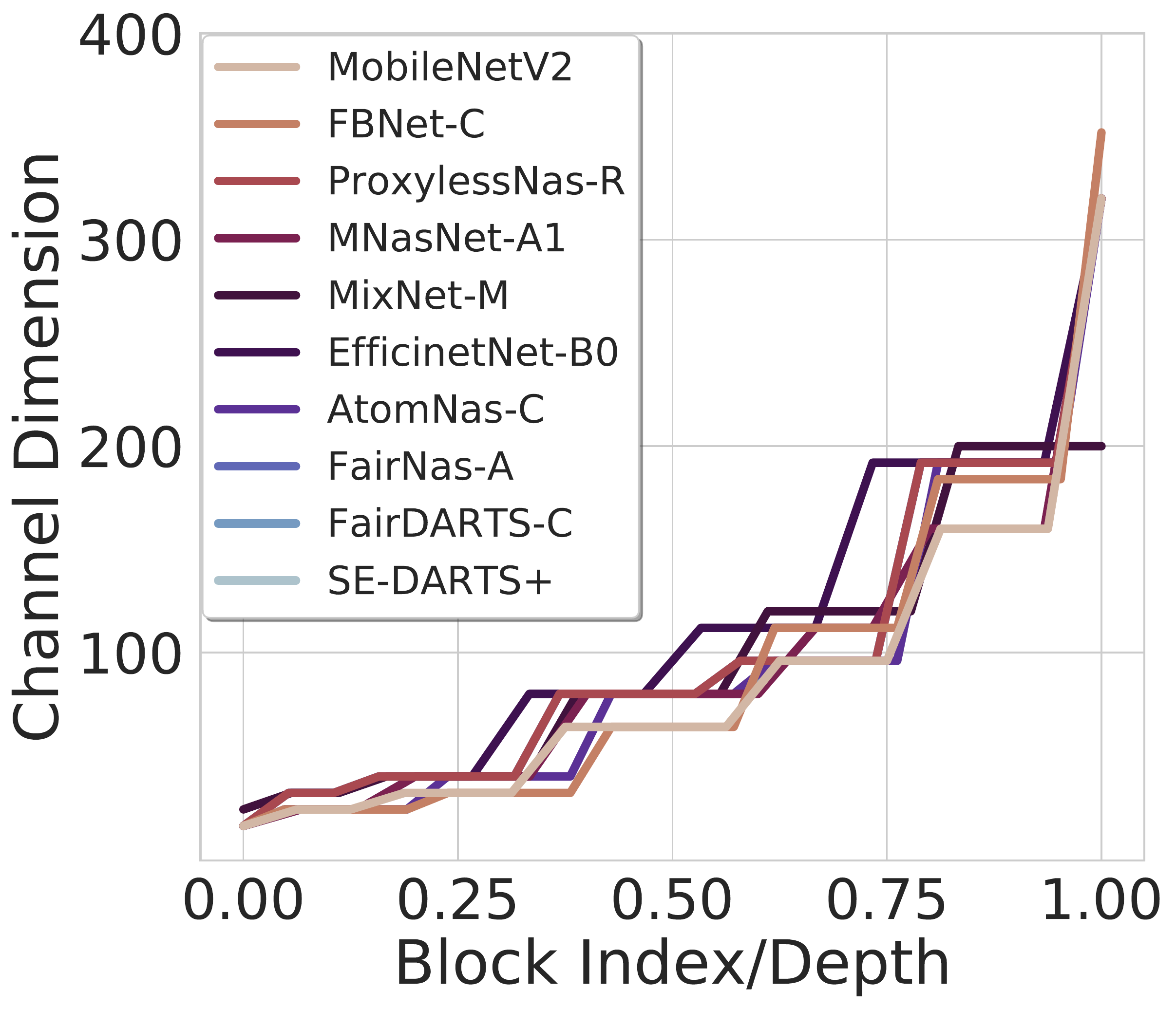}}
\vspace{-2mm}
\caption{\small {\bf Channel configurations in lightweight models.} We present the output channel dimensions of the stem's 3×3 convolution (e.g., 16, 24, 32) and the building blocks with the number of repeated layers (e.g., ×1, ×2, ×3, ×4, ×5) in block index order. All the models after MobileNetV2 have similar channel configurations to that of MobileNetV2, which have repeated channel dimensions for each stage.}%
\label{table:channel_dim_setting}
\vspace{-5mm}
\end{table*}

%% file: 02.Related_Work.tex
\section{Related Work}
\label{section:related}
After appearance of AlexNet~\cite{alexnet}, VGG~\cite{VGG}, GoogleNet~\cite{GoogleNet}, and ResNet~\cite{resnet} which show significant improvements in ImageNet classification, much lighter models such as~\cite{squeezenet, mobilenetv1,squeezenet} have been proposed with lowered computational budgets. Using the new operator depthwise convolution (dwconv)~\cite{mobilenetv1}, several architectures~\cite{xception, mobilenetv2, shufflenetv1, shufflenetv2} have been proposed with further efficient architecture designs. Taking advantage of the depthwise convolution could reduce a large amount of learnable parameters, and showed significant FLOPs reduction. Recently, structured network architecture search (NAS) methods have been proposed to yield the lightweight models~\cite{fbnet, chamnet, proxylessnas,mobilenetv3,mnasnet, tan2019mixconv,efficientnet, fbnetv2, mei2020atomnas, chu2019fairnas, chu2019fairdarts, liang2019dartsplus}, and EfficientNet~\cite{efficientnet} which based on compound scaling of width, depth, and resolution, became a de facto state-of-art model. Take one step forward from the existing lightweight models, we focus on finding an effective channel configuration for an inverted bottleneck module, which is an alternative to searching building blocks.%

%% file: 03-1.Preliminary.tex
\input{figures/figure__main_rank_test.tex}

\section{Designing an Expansion Layer}
\label{section:sec3}
In this section, we study how to design a layer properly considering the expressiveness, which is essential to design an entire network architecture.

\subsection{Preliminary}
\label{subsection:preliminary}
\paragraph{Estimating the expressiveness.} 
In language modeling, the authors~\cite{yang2017breaking} firstly highlighted that the softmax layer may suffer from turning the logits to the entire class probability due to the rank deficiency. This stems from the low input dimensionality of the final classifier and the vanished nonlinearity at the softmax layer when computing the log-probability. The authors proposed a remedy of enhancing the expressiveness, which improved the model accuracy. This implies that a network can be improved by dealing with the lack of expressiveness at certain layers. Estimating the expressiveness was studied in a model compression work~\cite{zhang2015efficient}. The authors compressed a model at layer-level by a low-rank approximation; investigated the amount of compression by computing the singular values of each feature. Inspired by the works, we conjecture that the rank may be closely related to the expressiveness of a network, and studying it may provide an effective layer design guide.
\vspace{-4mm}

\paragraph{Layer designs in practice.} ResNet families~\cite{resnet, preresnet,resnext} have bottleneck blocks doubling the input channel dimensions (i.e., 64-128-256-512 in order) to make the final dimension (2048) above the number of classes at last. The recent efficient models~\cite{fbnet, chamnet, proxylessnas,mobilenetv3,mnasnet, tan2019mixconv,efficientnet, fbnetv2, mei2020atomnas, chu2019fairnas, chu2019fairdarts, liang2019dartsplus} increase the channel dimensions steadily in inverted bottlenecks; therefore, they commonly involve a large expansion layer at the penultimate layer. The output dimension of the stem is set to 32 which expands the 3-dimensional input. Both of the inverted bottleneck~\cite{mobilenetv2} and bottleneck block~\cite{resnet} have the convolutional expansion layer with the predefined expansion ratio (mostly 6 or 4). Are these layers designed correctly and just need to design a new model accordingly? %

\vspace{-1mm}
\subsection{Empirical study}%
\label{subsection:sec_empirical_study}
\vspace{-1mm}
\paragraph{Sketch of the study.}
We aim to study a design guide of a single expansion layer that expands the input dimension. We measure the rank of the output features from the diverse architectures (over 1,000 random-sized networks) and see the trend as varying the input dimensions towards the output dimensions. The rank is originally bounded to the input dimension, but the subsequent nonlinear function will increase the rank above the input dimension~\cite{amini2011low, yang2017breaking}. However, a certain network fails to expand the rank close to the output dimension, and the feature will not be fully utilized. We statistically verify the way of avoiding failure when designing the network. The study further uncovers the effect of complicated nonlinear functions such as ELU~\cite{elu} or SiLU (Swish-1)~\cite{gelu,swish} and where to use them when designing lightweight models.

\vspace{-4mm}
\paragraph{Materials.}
We generate a bunch of networks with the building blocks consists of 1) a single 1×1 or 3×3 convolution layer; %
2) an inverted bottleneck block~\cite{mobilenetv2} with a 3×3 convolution or 3x3 depthwise convolution inside. We have the layer output (i.e., feature) $f(\mb{W}\mb{X})$ with $\mb{W}{\in}\mathbb{R}^{d_{out}\stimes d_{in}}$ and the input $\mb{X}{\in}\mathbb{R}^{d_{in}\stimes N}$, where $f$ denotes a nonlinear 
function\footnote{We use ReLU~\cite{relu}, ReLU6~\cite{mobilenetv2},
LeakyReLU~\cite{leakyrelu},
ELU~\cite{elu}, SoftPlus~\cite{softplus}, Hard Swish~\cite{mobilenetv3}, and SiLU (Swish-1)~\cite{gelu,swish}.} with the normalization (we use a BN~\cite{BN} here). $d_{out}$ is randomly sampled to realize a random-sized network, and $d_{in}$ is proportionally adjusted for each {\it channel dimension ratio} (i.e., $d_{in}/d_{out}$) in the range $[0.1, 1.0]$. $N$ denotes the batch size, where we set $N{>}d_{out}{>}d_{in}$.
We compute {\it rank ratio} (i.e., $\mathrm{rank}(f(\mb{W}\mb{X}))/d_{out}$) for each model and average them. An inverted bottleneck block is similarly handled as a single convolution layer to compute rank\footnote{We denote the input and output of an inverted bottleneck as the input of the first 1×1 convolution; the output after the addition operation of the shortcut and the bottleneck, respectively.}. 

\vspace{-4mm}
\paragraph{Observations.} Figure~\ref{fig:rank_test} shows how the rank changes with respect to the input channel dimension on average. Dimension ratio ($d_{in}/d_{out}$) on x-axis denotes the reciprocal of the expansion ratio~\cite{mobilenetv2}. We observe the followings: 
\vspace{-2mm}
\begin{enumerate}[label=(\roman*)]
    \item {\it Drastic channel expansion harms the rank.} This holds for a single convolution layer and an inverted bottleneck both as shown in Figure~\ref{fig:rank_test}. The impact gets bigger with 1) an inverted bottleneck (see Figure~\ref{subfig:IB_conv}); 2) a depthwise convolution (see Figure \ref{subfig:IB_dwconv})\footnote{Actually, many recent lightweight models avoid increasing the channel dimension drastically (i.e., dimension ratio is usually higher than 0.5) at an inverted bottleneck; the models have inverted bottlenecks where the first 1×1 convolution has the expansion ratio 3, 4, or 6. Figure~\ref{subfig:layer1x1} tells us the design factors are tenable; the architects may know this empirically, but we believe that we have provided an underlying reason through the study.};
    \vspace{-2mm}
    \item {\it Nonlinearities expand rank.} Figure~\ref{fig:rank_test} shows averaged rank is expanded above the input channel dimension by nonlinear functions. They expand rank more at smaller dimension ratio, and complicated ones such as ELU, or SiLU do more;
    \vspace{-2mm}
    \item {\it Nonlinearities are critical for convolutions.} Nonlinearities expand the rank of 1×1 and 3×3 single convolutions more than an inverted bottleneck (see Figure~\ref{subfig:layer1x1} and \ref{subfig:layer3x3} vs. Figure~\ref{subfig:IB_conv} and \ref{subfig:IB_dwconv}). %
    \vspace{-4mm}
\end{enumerate}

\paragraph{What we learn from the observations.}
We learned the followings: 1) an inverted bottleneck is needed to design with the expansion ratio of 6 or smaller values at the first 1×1 convolution; 2) each inverted bottleneck with a depthwise convolution in a lightweight model needs a higher channel dimension ratio; 3) a complicated nonlinearity such as ELU and SiLU needs to be placed after 1×1 convolutions or 3×3 convolutions (not depthwise convolutions). Based on the knowledge, in the following section, we perform channel dimension searches to find an effective channel configuration for entire channel dimensions. This is to uncover whether the conventional way of configuring channels shown in Table~\ref{table:channel_dim_setting} is optimal or not, albeit the models have worked well with high accuracy.

\input{tables/toy_exp1} 
\vspace{-4mm}
\paragraph{Verification of the study.} We finally provide an experimental backup to make sure what we have learned contributes to improving accuracy. We train the models consisting of two inverted bottlenecks (IBs)\footnote{The models with a single IB cannot be similar in computational complexity with the fixed stem, so the models should contain at least two IBs.} to adjust the channel dimension ratio (DR) of IBs and the first 1×1 convolutions in each IB. Starting from the baseline with the low DR 1/20, we successively study through the picked models with 1) increasing DR of the first 1×1 conv to 1/6; 2) increasing DR at every IB from 0.22 to 0.8; 3) replacing the first ReLU6 with SiLU in each IB. Table~\ref{table:toy_exp1} shows each factor works well, and the rank and accuracy increase together.

%% file: figures/figure__main_rank_test.tex
\begin{figure*}[t]
\small
\centering
\hspace{-4mm}
\begin{subfigure}[ht!]{0.24\linewidth}
\includegraphics[trim = 0mm 0mm 0mm 0mm, clip, width=1.0\linewidth]{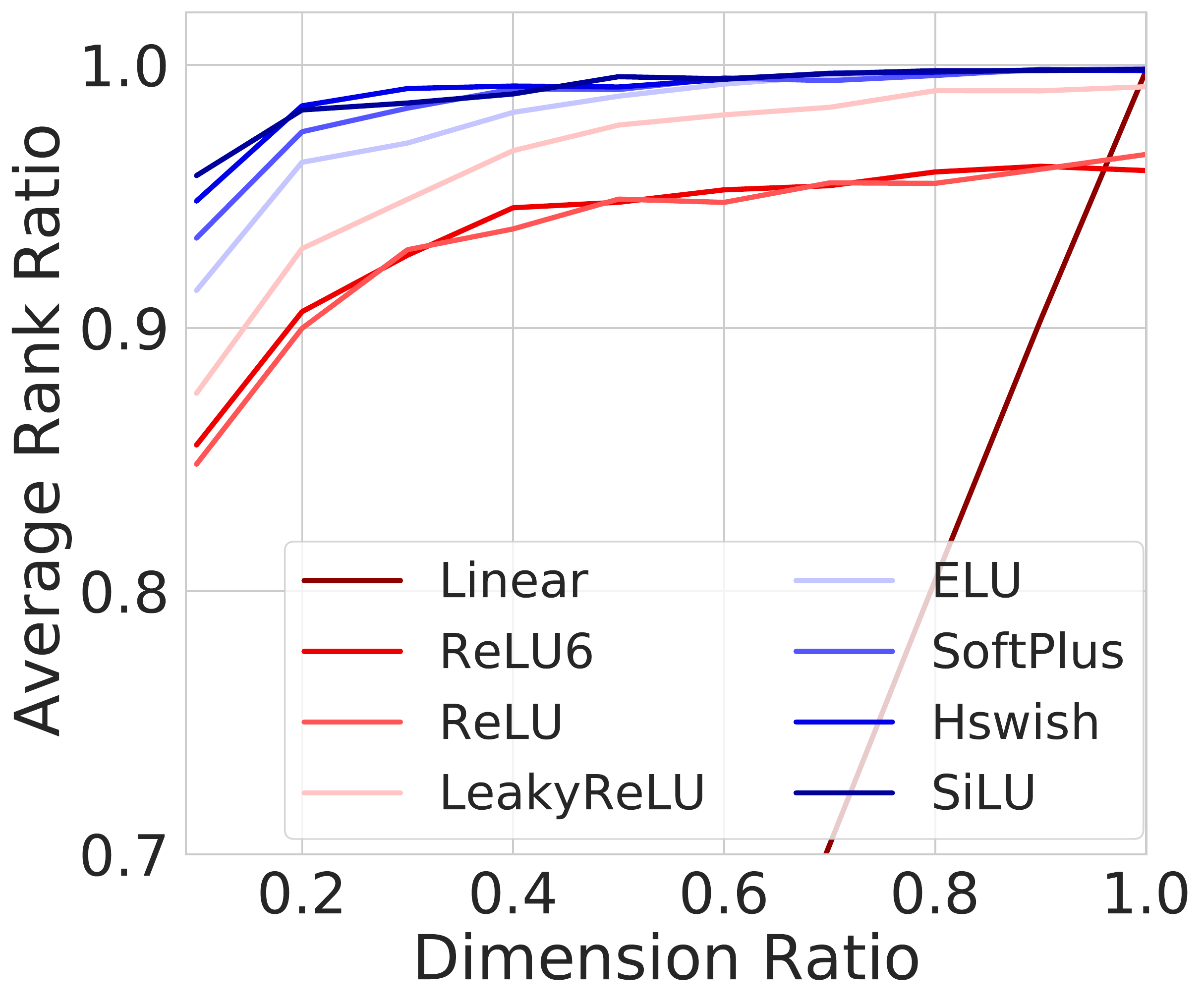}
\vspace{-6mm}
\caption{\small 1×1 convolution}
\label{subfig:layer1x1}
\end{subfigure}
\begin{subfigure}[ht!]{0.24\linewidth}
\includegraphics[trim = 0mm 0mm 0mm 0mm, clip, width=1.0\linewidth]{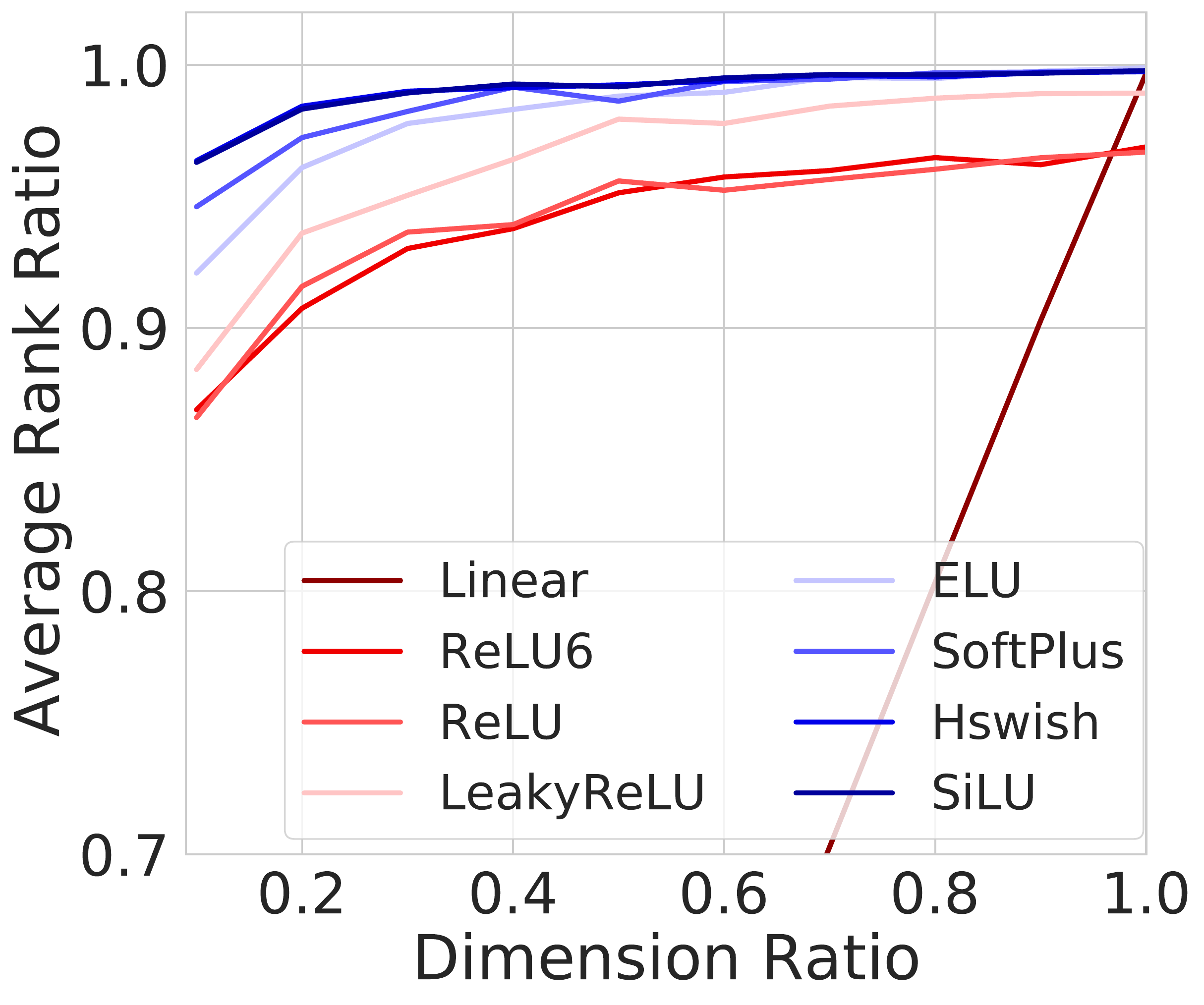}
\vspace{-6mm}
\caption{\small 3×3 convolution}
\label{subfig:layer3x3}
\end{subfigure}
\begin{subfigure}[ht!]{0.24\linewidth}
\includegraphics[trim = 0mm 0mm 0mm 0mm, clip, width=1.0\linewidth]{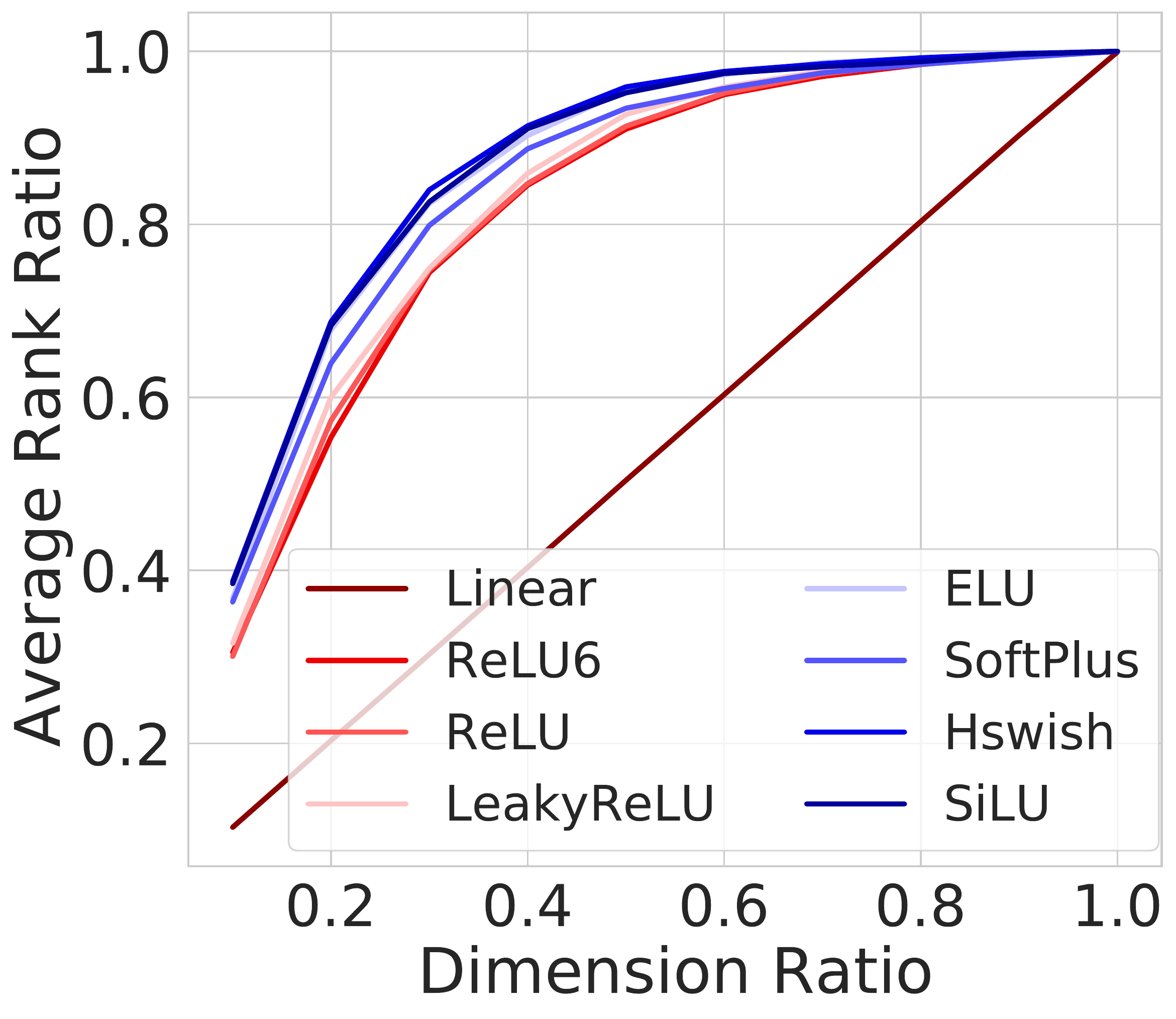}
\vspace{-6mm}
\caption{\small Inverted Bot. (with conv)}
\label{subfig:IB_conv}
\end{subfigure}
\begin{subfigure}[ht!]{0.24\linewidth}
\includegraphics[trim = 0mm 0mm 0mm 0mm, clip, width=1.0\linewidth]{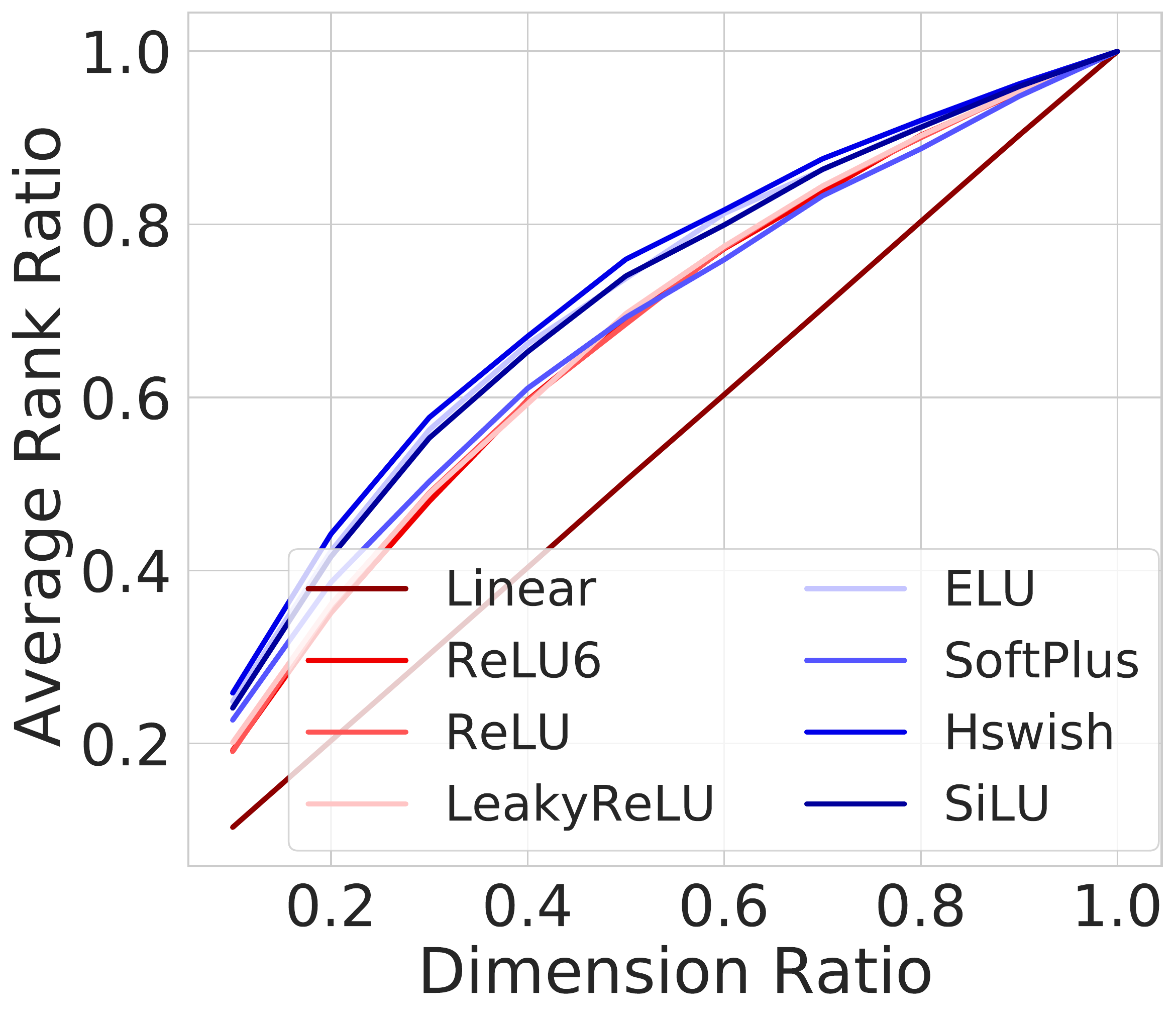}
\vspace{-6mm}
\caption{\small Inverted Bot. (with dwconv)}
\label{subfig:IB_dwconv}
\end{subfigure}
\vspace{-3mm}
\caption{\small {\bf Visualization of the output rank}. We measure the rank ratio (i.e., rank/output channel dimension) vs. channel dimension ratio (i.e., input channel dimension/output channel dimension) from diverse architectures averaged over 1,000 random-sized networks with various nonlinear functions: (a) A single 1×1 convolution; (b) A single 3×3 convolution; (c) An inverted bottleneck with a 3×3 convolution; (d) An inverted bottleneck with a 3×3 depthwise convolution~\cite{mobilenetv2}. We fundamentally observe that all the ranks are expanded above the input channel dimensions by the nonlinear functions with different network architectures.}
\label{fig:rank_test}
\vspace{-5mm}
\end{figure*}

%% file: tables/toy_exp1.tex
\begin{table}[t]
\fontsize{8.5}{9.5}\selectfont
\centering
\tabcolsep=0.1cm
\hspace{-1.25mm}
\begin{tabular}{@{}l|c|c|c@{}}
Network & FLOPs & Top-1 & Nuc. norm \\
\midrule
Baseline  & 103M & 48.0\% & 5997.5  \\ 
+ Increase DR of 1×1 conv ($1/20{\rightarrow}1/6$) & 99M & 52.1\% & 6655.9\\
+ Increase DR of IB ($0.22{\rightarrow}0.8$) & 105M & 53.8\% & 6703.2\\ 
+ Replace ReLU6 with SiLU & 105M & 54.6\%  & 6895.9\\
\end{tabular}
\quad
\vspace{-3mm}
\caption{\small {\bf Factor analysis of the study}. We use four different models with similar computational complexity and report the accuracy and rank represented by the nuclear norm of the final feature. We average all the numbers over three models trained on CIFAR-100~\cite{cifar}. Each factor successively improves the accuracy and expands the rank without extra computational cost.}
\label{table:toy_exp1}
\vspace{-5mm}
\end{table}

%% file: 03-2.Problem_Formulation.tex
\input{figures/figure__searched_model_width.tex}

\input{tables/channel_search_result.tex}

\section{Designing with Channel Configuration}
\label{section:sec4}

\subsection{Problem Formulation}
\vspace{-1mm}
Our goal is to reveal an effective channel configuration of designing a network under the computational demands. We formulate the following problem:
\begin{align}
     \max_{c_i, i=1 \dots d} &  \text{Acc}({\it N}(c_1, \dots c_d)) \nonumber  \\ 
      \text{s.t.\quad} & c_1   \leq c_2\leq \dots \leq c_{d-1} \leq c_d, \nonumber  \\
      & \text{Params}(N)  \leq P, \ \ \text{FLOPs}(N) \leq F,  %
      \label{eq:problem}
\end{align}
where \text{Acc} denotes the top-1 accuracy of the model; $c_i$ denotes output channel of $i$-th block among $d$ building blocks; $P$ and $F$ denote the target parameter size and FLOPs. We involve the monotonic increasing of $c_i$ because this contains the channel configurations shown in Table~\ref{table:channel_dim_setting}; we note that the opposite case, channel dimension consistently decreasing, requires a hard-to-use amount of FLOPs. Here, we are not targeting a hardware-specific model, so we concern with FLOPs rather than inference latency. Notice that many NAS methods~\cite{fbnet, proxylessnas,mobilenetv3,mnasnet, tan2019mixconv,efficientnet,mei2020atomnas, chu2019fairnas, chu2019fairdarts, liang2019dartsplus} search the network $N$ with fixing $c_i$ with based on the predefined channel configurations as shown in Table~\ref{table:channel_dim_setting}, but on the other hand we search $c_i$ while fixing the network $N$.%

\subsection{Searching with channel parameterization}
\vspace{-1mm}
We observe a general trend of architectural shapes for diverse computational demands through searches. Alternative to optimizing eq.\eqref{eq:problem} directly, we represent the channel dimensions at each building block with a piecewise linear function, which can reduce the search space. We parameterize the channel dimensions as $c_i=af(i)+b$, where $a$ and $b$ are to be searched; let $f(i)$ as a piecewise linear function by picking a subset of $f(i)$ up from ${1\dots d}$. In this fashion, the channel parameterizations contain the conventional MobileNetV2-based channel configurations (i.e., stage-wise channel configurations) shown in Table~\ref{table:channel_dim_setting}. We adopt the CIFAR-10 and CIFAR-100 datasets~\cite{cifar} as done in NAS methods~\cite{darts,chen2019pdarts,liang2019dartsplus} to search the parameterization.

To control the other variables, we set all the networks that have the fixed channel dimension at the stem 3×3 convolution of 16 followed by a BN~\cite{BN} with a ReLU and have the large expansion layer at the penultimate layer. We use the original inverted bottleneck (expansion ratio of 6)~\cite{mobilenetv2} as our building blocks, which is a fundamental block of lightweight NAS methods, so we do not search the building blocks' expansion ratio. The chosen elements are based on the above investigation of single-layer design. 
Optimization is done alternatively by searching and training a network. We train each model for 30 epochs for faster training~\cite{regnet} and the early stopping strategy~\cite{liang2019dartsplus}. Each training is repeated three times for averaging the accuracies to reduce the accuracy fluctuation caused by random initialization. 

\subsection{Search Results}
\vspace{-1mm}
We perform individual searches under different constraints of computational costs to provide reliable and generalizable search results. We assign four search constraints to aim to search across different target model sizes. After each search, we collect top-10\%, middle-10\% (i.e., the models between top-50\% and 60\%), and bottom-10\% models in terms of the model accuracy from 200 searched models to analyze them. To this end, we first visualize the collected models' channel configuration in Figure~\ref{fig:searched_model_stats} of each search; we then report the detailed performance statistics of the models and the best/worst-performing models' channel configuration in Table~\ref{table:channel_search_result}.

Figure~\ref{fig:searched_model_stats} illustrates the clear trends in which channel configuration is more effective in terms of accuracy. We observe that the {\it linear parameterizations} by the block index as colored with {\color{red} red} enjoy higher accuracies while maintaining similar computational costs. This parameterization is regularly found throughout searching in diverse environments as shown in the figure. Note that the best models in Table~\ref{table:channel_search_result} have the channels configured with linearly increasing that is identical to the linear parameterization. The models in {\color{ForestGreen} green} have highly reduced the input-side channels and therefore, most of the weight parameters are placed at the output-side resulting in the loss of accuracy. In addition, intriguingly, {\color{RoyalBlue} blue} represents the models at the middle 10\%-accuracy, which looks similar to the channel configurations in Table~\ref{table:channel_dim_setting}.
Their conventional configurations are designed to attain a flop-efficiency by limiting the channels at earlier layers and giving more channels close to the output. Therefore, we may safely suggest that we need to change the convention ({\color{RoyalBlue} blue}) towards the new channel configuration ({\color{red} red}) that can achieve additional accuracy gains. %

We have additional search results shown in Figure~\ref{fig:depthfixed_searched_model_stats} when fixing the network depth and searching under the different computational costs; we found identical linear parameterizations. Furthermore, we perform the same experiment on CIFAR-10 with the constraints of computational costs and found the same trends in Figure~A1. %
As mentioned earlier, we believe that the success of the works~\cite{pyramidnet,densenet} may stem from similar parameterizations. In the case of training the 18-depth models, which have about 30 MFLOPs, it takes 1.5 GPU days for searching and training about 100 models for individual 30 epochs training with 3 runs for training each model. The training cost for the entire search is $30\,\text{MFLOPs}\stimes30\,\text{epochs}\stimes3\,\text{runs}\stimes100\,\text{models}=2.7\stimes100\,\text{GFLOPs}$ epochs, and this is smaller than training a single ImageNet model ResNet50 for 100 epochs ($4.0\,\text{GFLOPs}\stimes100\,\text{epochs}$). Although the actual training time is less than that of ResNet50, comparing directly with the ImageNet model may be inappropriate. We provide a rough estimate of the amount of computation in practice.

%% file: figures/figure__searched_model_width.tex
\begin{figure*}[t]
\small
\centering
\hspace{-2mm}
\begin{subfigure}[ht!]{0.24\linewidth}
\includegraphics[page=1, trim = 0mm 0mm 0mm 0mm, clip, width=1.0\linewidth]{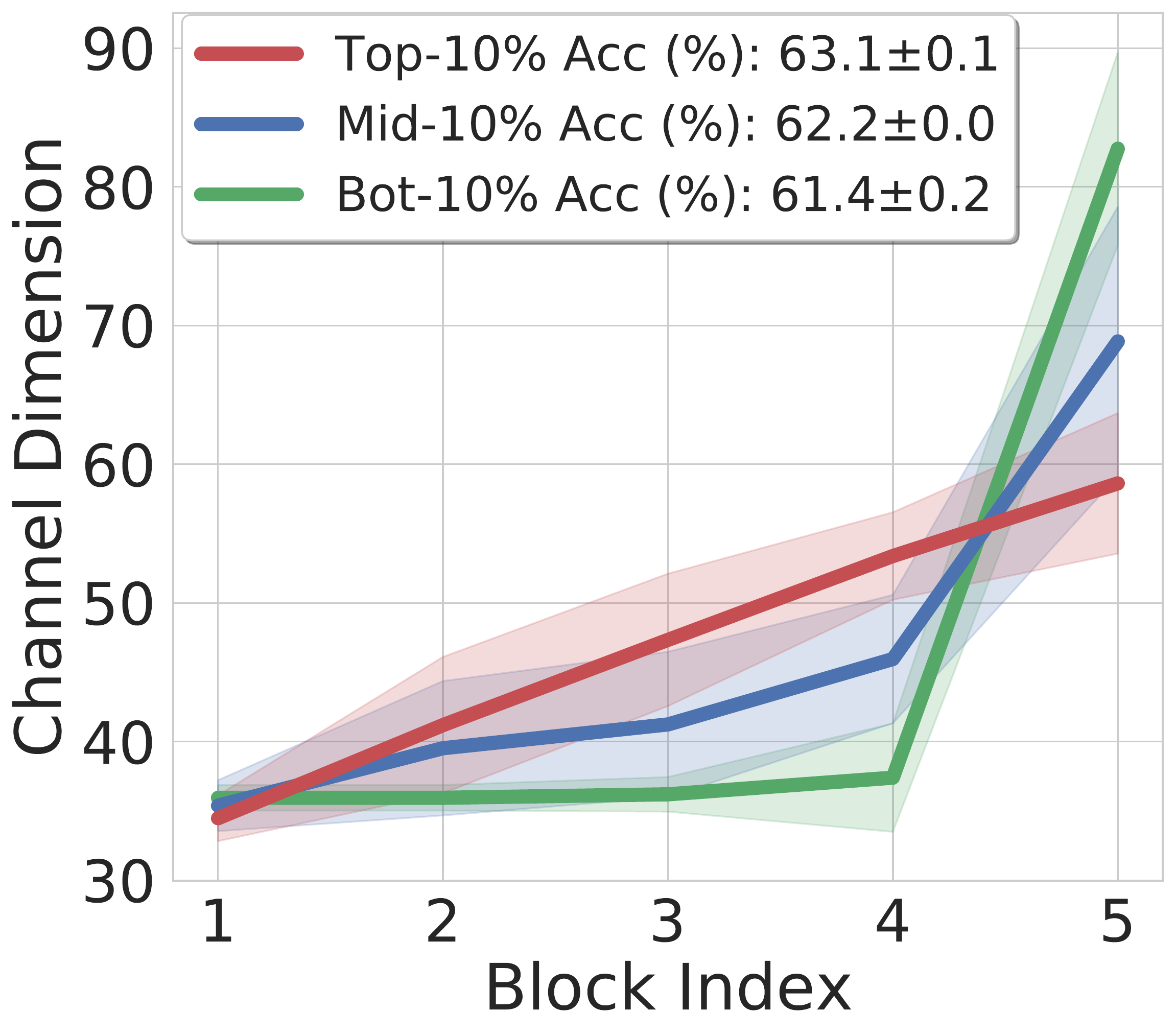} 
\vspace{-6mm}
\caption{\small 18-depth models}
\end{subfigure}
\begin{subfigure}[ht!]{0.24\linewidth}
\includegraphics[page=1, trim = 0mm 0mm 0mm 0mm, clip, width=1.0\linewidth]{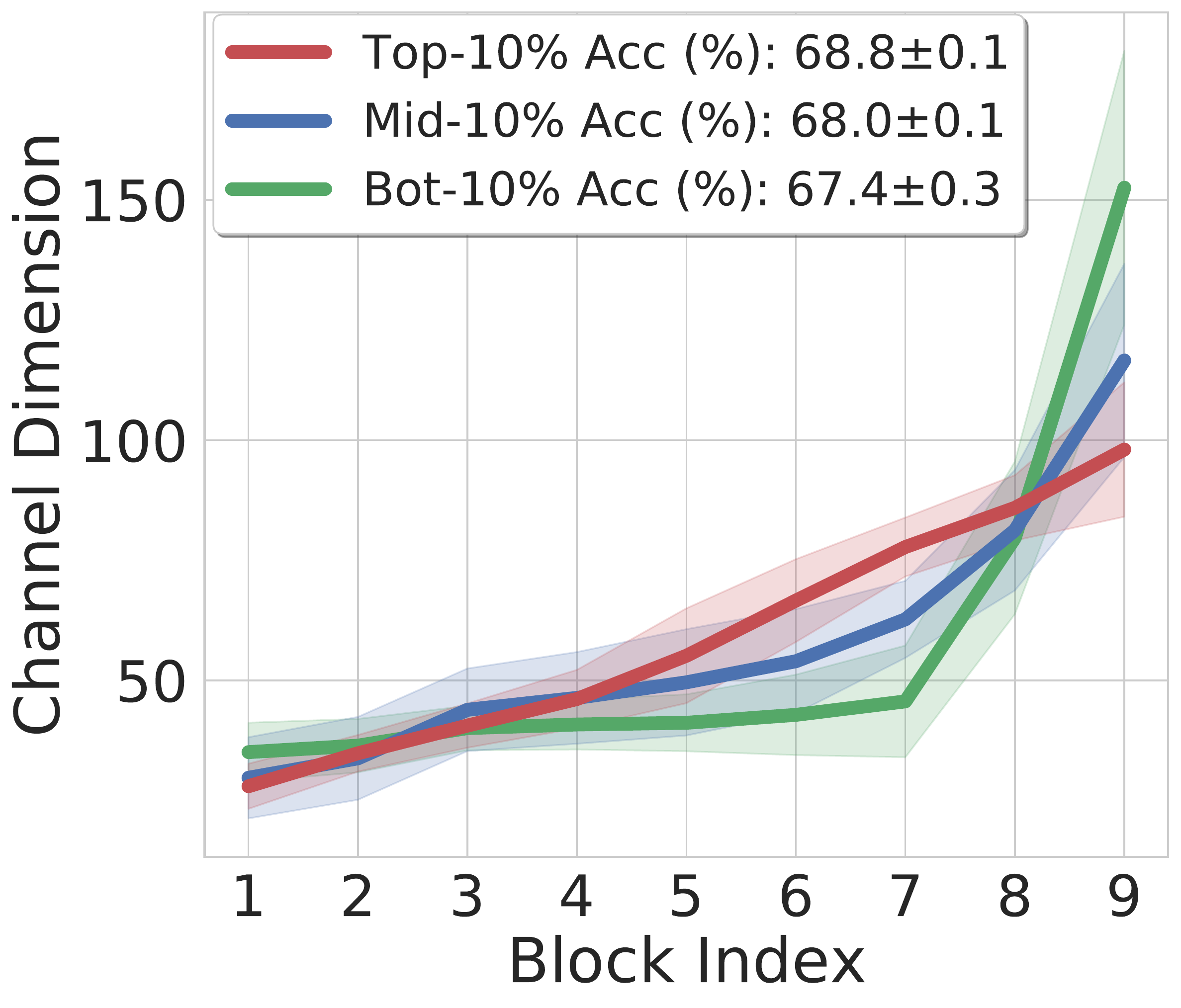} 
\vspace{-6mm}
\caption{\small 30-depth models}
\end{subfigure}
\begin{subfigure}[ht!]{0.24\linewidth}
\includegraphics[page=1, trim = 0mm 0mm 0mm 0mm, clip, width=1.0\linewidth]{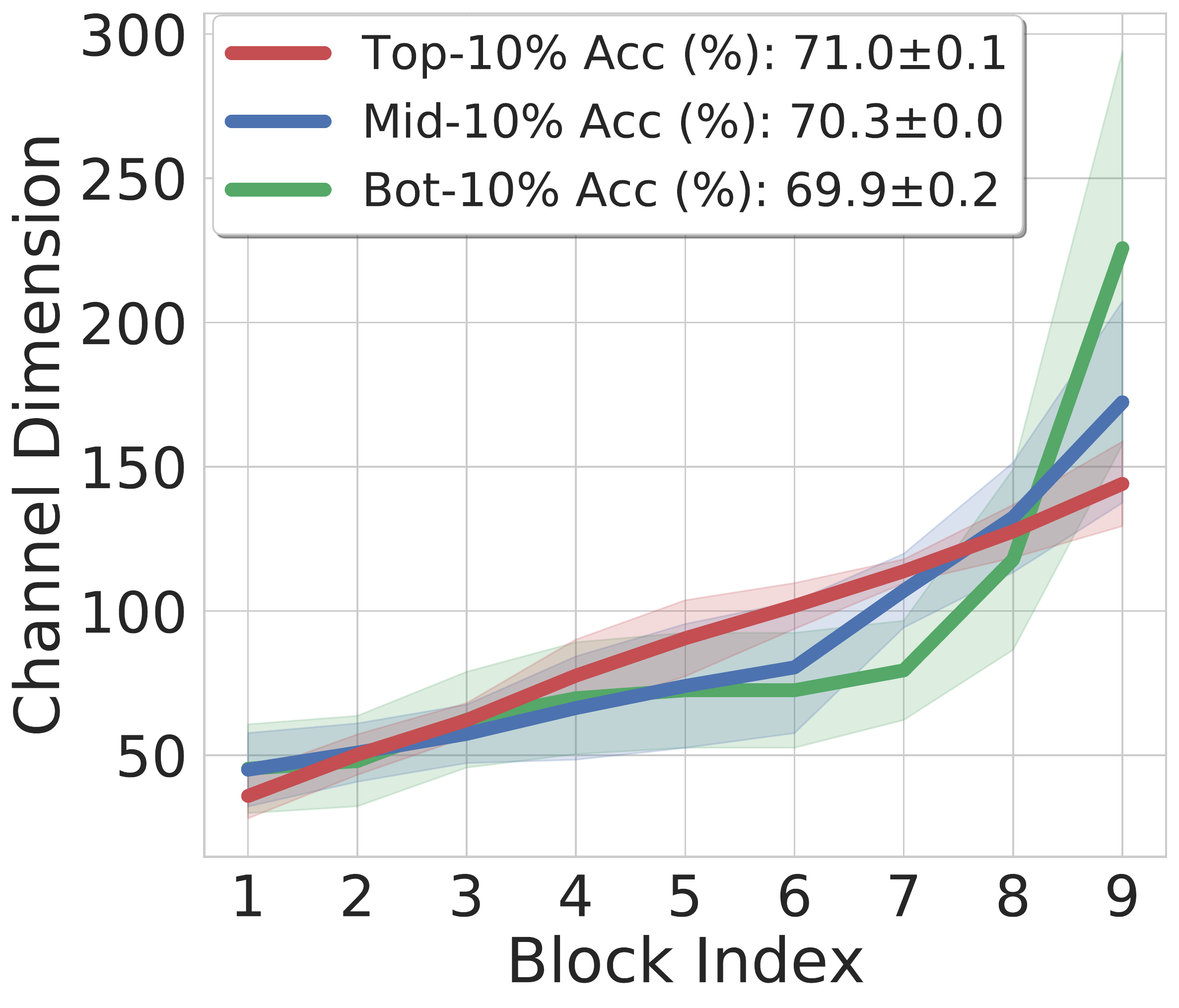} 
\vspace{-6mm}
\caption{\small 30-depth models}
\end{subfigure}
\begin{subfigure}[ht!]{0.24\linewidth}
\includegraphics[page=1, trim = 0mm 0mm 0mm 0mm, clip, width=1.0\linewidth]{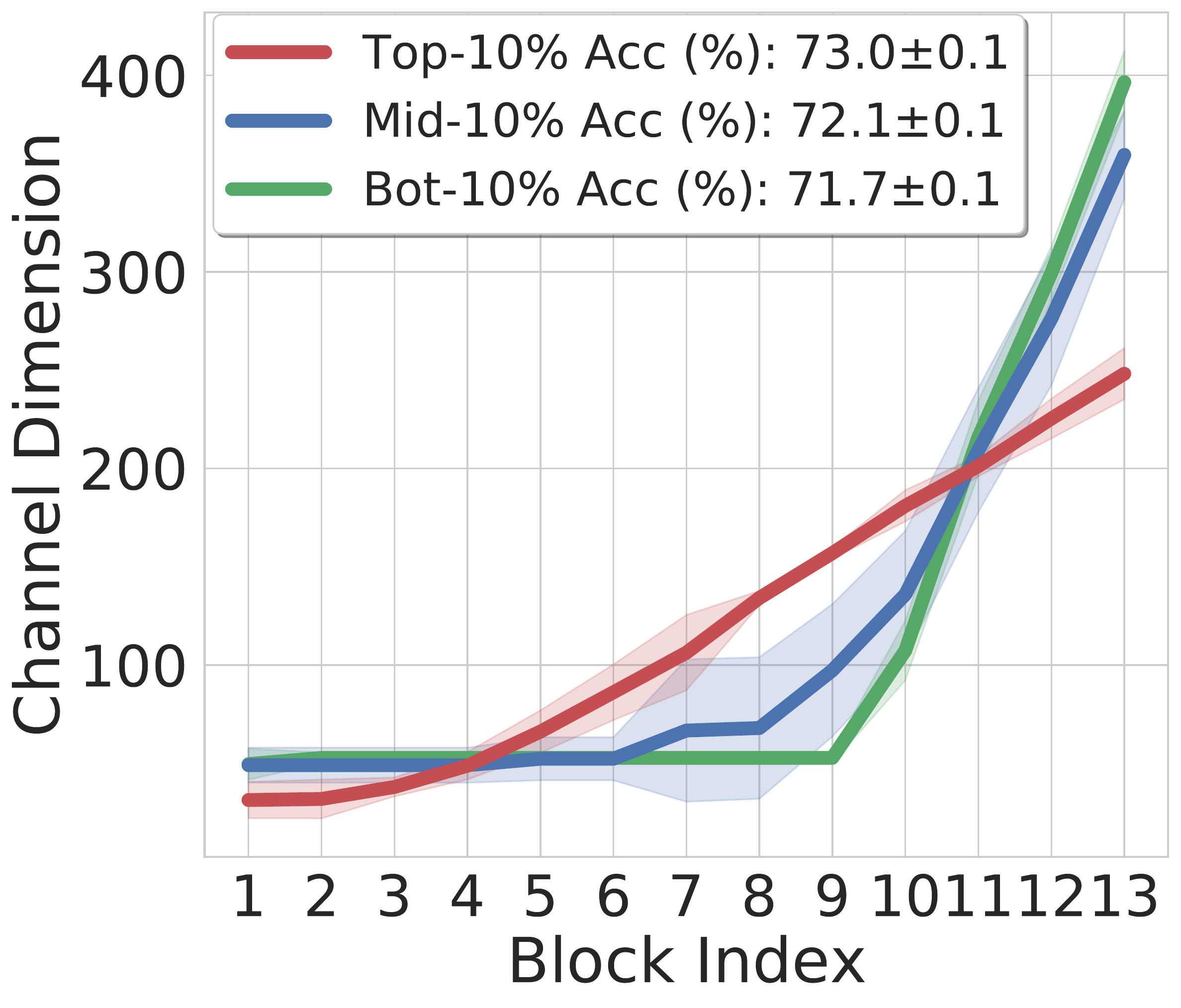} 
\vspace{-6mm}
\caption{\small 42-depth models}
\end{subfigure}
\vspace{-3mm}
\caption{\small {\bf Visualization of the searched models' channel dimensions vs. block index}. {\color{red} Red}: top-10\%; {\color{RoyalBlue} blue:} middle-10\%; {\color{ForestGreen} green}: bottom-10\% accuracy models; we plot the averaged channel configurations with the 1-sigma range over each searched candidate.}%
\label{fig:searched_model_stats}
\vspace{-2mm}
\end{figure*}

%% file: tables/channel_search_result.tex
\begin{table*}[t]
\fontsize{7.8}{8.2}\selectfont
\centering
\tabcolsep=0.14cm
\renewcommand{\arraystretch}{0.5}
\begin{tabular}{@{}l|l|c|c|c|l@{}}
Search constraints & Ranking & Acc (\%) & Params (M) & FLOPs (M) & Best and worst models among searched configurations \\
\midrule
\multirow{3}{*}{\shortstack[l]{(a) Models with 5 inverted bot., \\ \# Params$\simeq$0.2M, FLOPs$\simeq$30M}} 
                  & Top-10\%     & 63.1±0.1 & 0.2±0.0  & 30±1 & \multirow{3}{*}{\shortstack[l]{Best: 34-34-45-55-66 (Acc: 63.4\%)\\ Worst: 36-36-36-36-83 (Acc: 61.1\%)}} \\ 
                  & Mid-10\%  & 62.2±0.0 & 0.2±0.0  & 30±1  &  \\
                  & Bot-10\%  & 61.4±0.2 & 0.2±0.0  & 30±1  &  \\ 
\midrule
\multirow{3}{*}{\shortstack[l]{(b) Models with 9 inverted bot.,\\ \# Params$\simeq$0.5M, FLOPs$\simeq$100M}} 
                  & Top-10\%     & 68.8±0.1 & 0.5±0.0 & 101±5 & \multirow{3}{*}{\shortstack[l]{Best: 24-33-42-50-59-68-77-85-94 (Acc: 68.9\%) \\ Worst: 39-39-39-39-39-39-39-87-158 (Acc: 67.6\%)}} \\
                  & Mid-10\%  & 68.0±0.1 & 0.5±0.0 & 100±6   &  \\
                  & Bot-10\%  & 67.4±0.3 & 0.5±0.0  & 97±8 &   \\
\midrule
\multirow{3}{*}{\shortstack[l]{(c) Models with 9 inverted bot.,\\ \# Params$\simeq$1.0M, FLOPs$\simeq$200M}} 
                  & Top-10\%     & 71.0±0.1 & 1.0±0.0  & 210±15  & \multirow{3}{*}{\shortstack[l]{Best: 30-45-59-74-88-103-117-132-146 (Acc: 71.1\%)\\ Worst: 47-47-70-70-70-70-70-70-364 (Acc: 69.6\%)}}\\
                  & Mid-10\%  & 70.3±0.0 & 1.0±0.0  & 198±18  &  \\
                  & Bot-10\%  & 69.9±0.2 & 1.0±0.0  & 200±15  &  \\
\midrule
\multirow{3}{*}{\shortstack[l]{(d) Models with 13 inverted bot.,\\ \# Params$\simeq$3.0M, FLOPs$\simeq$300M}}
                  & Top-10\%     & 73.0±0.1 & 3.0±0.0  & 351±5  & \multirow{3}{*}{\shortstack[l]{Best: 34-34-34-40-64-88-112-136-160-184-208-232-256 (Acc: 73.2\%) \\ Worst: 52-52-52-52-52-52-52-52-52-115-263-263-412 (Acc: 71.6\%)}} \\
                  & Mid-10\%  & 72.1±0.1 & 3.0±0.0  & 351±5  &  \\
                  & Bot-10\%  & 71.7±0.1 & 3.0±0.0  & 351±6  &  \\
\midrule
            
\end{tabular}
\vspace{-4mm}
\caption{\small {\bf Detailed searched channel configurations.} We provide the detailed individual searched results under the different search constraints. Along with Figure~\ref{fig:searched_model_stats}, we report the detailed numbers including the averaged accuracy, \# of parameters, FLOPs over top-10\%, midddle-10\%, and bottom-10\% accuracy models. We further present each of the best and the worst models' channel configurations.}
\label{table:channel_search_result}
\vspace{-5mm}
\end{table*}

%% file: 04.Network_Improvement.tex
\subsection{Network upgrade}
\label{subsection:network_upgrade}
\vspace{-1mm}
We rebuild the existing model based on the investigations in practice. From the baseline MobileNetV2~\cite{mobilenetv2} which introduced the convention of channel configuration, we only reassign each output channel dimension of inverted bottlenecks by following the parameterization. We use the identical setting of the stem (i.e., 3×3 convolution with BN and ReLU6) and the inverted bottleneck with the expansion ratio 6. We use the same large expansion layer at the penultimate layer. To fairly compare with the aforementioned lightweight models MobileNetV3~\cite{mobilenetv3}, MixNet~\cite{tan2019mixconv}, EfficientNet~\cite{efficientnet}, AtomNas~\cite{mei2020atomnas}, FairNAS~\cite{chu2019fairnas}, FairDARTS~\cite{chu2019fairdarts}, DART+~~\cite{liang2019dartsplus}, we further replace ReLU6 with SiLU~\cite{gelu,swish} and adopt SE~\cite{SENet} in the inverted bottlenecks. 

\vspace{-0.5mm}

Based on the investigation in \S\ref{section:sec3}, we replace ReLU6 only after the first 1×1 convolution in each inverted bottleneck because we observed the layer with a smaller dimension ratio needs to be more addressed; the second depthwise convolution has the channel dimension ratio of 1, so we do not replace ReLU6 here. This can further realize the simplicity of model design and benefit from faster training speeds since ReLU6s remain. Using other nonlinear functions such as ELU shows similar accuracy gains (see Table~A2, %
but we use SiLU (Swish-1) for a fair comparison with the aforementioned lightweight models which use SiLU. 

\vspace{-0.5mm}

Note that only with these simple modifications, our model outperforms NAS-based methods in many experiments (\S\ref{section:exp}), which signifies the importance of the channel configuration. We call our model Rank Expansion Networks (ReXNet) as observed the actual rank expansion in \S\ref{section:ablation_and_discussion}. Additionally, we build another model with the linear parameterization upon MobileNetV1~\cite{mobilenetv1}, which shows a large accuracy improvement (+2.3pp) over the baseline 72.5\%. We call this model ReXNet (plain). The detailed model specification of ReXNets is provided in Appendix~\ref{app:models}. %

%% file: 05.Experiment.tex
\section{Experiment}
\label{section:exp}
\input{tables/imagenet_vs_light.tex}

\subsection{ImageNet Classification}
\label{subsection:imagenet_classification}
\vspace{-1mm}
\paragraph{Training on ImageNet.} We train our model on the ImageNet dataset~\cite{imagenet} using the standard data augmentation~\cite{GoogleNet} with stochastic gradient descent (SGD) and mini-batch size of 512 on four GPUs. Learning rate is initially set to 0.5 and is scheduled by cosine learning rate scheduling. Weight decay is set to 1e-5. Table~\ref{table:imagenet_vs_light} show the performance comparison with popular lightweight models where all the reported models are trained and evaluated with $224\stimes224$ image size. %
Comparing with the models~\cite{mobilenetv3,mnasnet,tan2019mixconv,efficientnet} with SE~\cite{SENet} and SiLU\footnote{MobileNetV3 uses Hard Swish~\cite{mobilenetv3}, and MnasNet does not use SiLU.}~\cite{gelu,swish}, our model outperforms the most of the models searched by NAS including MobileNetV3-Large, MNasNet-A3, MixNet-M, and EfficientNet-B0 (without AutoAug~\cite{autoaug}) with at least {\bf +0.3pp} accuracy improvement. Our model outperforms MixNet-M and AtomNas-C+ which use mixed-kernel operations under similar computational costs.

Additionally, we train our model with RandAug~\cite{cubuk2019randaugment} to compare it fairly with the NAS-based models~\cite{efficientnet,chu2019fairnas,chu2019fairdarts,liang2019dartsplus} using additional regularizations (denoted by $^*$ in Table~\ref{table:imagenet_vs_light}) such as Mixup~\cite{mixup}, AutoAug~\cite{autoaug}, and RandAug~\cite{cubuk2019randaugment}, our model improves all the models including EfficientNet-B0 (with AutoAug), FairNas-A, FairDARTS-C, and SE-DARTS+ by at least {\bf +0.4pp}\footnote{ReXNet (×1.0) has been improved to {\bf 78.1}\% trained by the novel optimizer AdamP~\cite{heo2020adamp} replacing SGD and further improved to {\bf 78.4}\% trained with the new training method ReLabel~\cite{yun2021re}, respectively.}. %
Strikingly, our model does not require further searches, but it either outperforms or is comparable to NAS-based models.

\input{figures/figure__vs_efficientnets.tex}
\input{tables/imagenet_vs_eff.tex}
\input{tables/coco_overall.tex}

\input{tables/coco_frcnn.tex}
\input{tables/transferlearning.tex}

\vspace{-4mm}
\paragraph{Comparison with Efficientnets.} We compare ReXNets with EfficientNets~\cite{efficientnet} about model scalability with the performances. To this end, we adopt the width multiplier concept~\cite{mobilenetv1, mobilenetv2, shufflenetv1, shufflenetv2, mobilenetv3} for scaling. Note that all our models are trained with the fixed resolution $224\stimes224$ unlike EfficientNets trained with the resolutions from $224\stimes224$ to $600\stimes600$. We do not use the method such as FixResNet~\cite{fixresnet} which performs additional training.

Table~\ref{table:imagenet_vs_efficientnet} shows our models adjusted by the multiplier from ×0.9 to ×2.0 with remarkable accuracy increments (see more models in Appendix~\ref{app_sub:model_models}). 
We measure CPU and GPU inference speeds to show the efficiency; we average the latencies over 1,000 runs with the batch size 1 on an Intel Xeon CPU E5-2630 and the batch size 64 on a V100 GPU, respectively. Figure~\ref{fig:vs_efficinetnets} visualizes our models' computational efficiency compared with EfficientNets; as the model size is larger, we observe that our models are much faster than EfficientNets. Notice that ReXNet (×2.0) is about {\bf 1.4× and 2.0× faster} than EfficientNet-B3 on CPU and GPU, respectively with almost the same accuracy and improves {\bf +2.5pp} top-1 accuracy on EfficientNet-B1 at a similar speed. This benefit may come from the fixed resolution and network depth for training and inference over all the models which can reduce the memory access time.

\input{tables/coco_maskrcnn.tex}

\subsection{COCO object detection}
\vspace{-1mm}
\label{subsection:coco_detection}
\paragraph{Training SSDLite.} 
We validate our backbones through object detection on the COCO dataset~\cite{coco2017} in SSDLite~\cite{mobilenetv2} which has lightweight detection heads suitable for seeing the feature extractor's capability. We follow the identical design elements~\cite{mobilenetv2,mobilenetv3,mnasnet} by building the first head on top of the final expansion layer which has the stride 16 and another head on top of the final layer. We train ReXNets (×0.9, ×1.0, and ×1.3) and EfficientNets-B0, B1, and B2 in SSDLites with the same training settings~\cite{mobilenetv2,mobilenetv3,mnasnet} including $320\stimes320$ image resolution for fair comparison.%

Table~\ref{table:coco_overall} shows ours largely outperform the other backbones with comparable computational costs. ReXNet (×1.0) outperforms EfficientNet-B1 trained with $240\stimes240$ by {\bf +1.8pp}, and ReXNet (×1.3) outperforms EfficientNet-B2 trained with $260\stimes260$ by {\bf +0.7pp} under similar computational costs. Interestingly, ReXNet (×0.9) achieves {\bf +1.2pp} AP improvement over EfficientNet-B0 with less computational costs. The large AP improvements indicate our channel dimension configuration can help finetuning as well. We provide more detection results in Appendix~\ref{app:coco_detection}. %
Note that ReXNet (×0.9) is faster than EfficientNet-B0 (75ms vs. 77ms), and ReXNet (×1.3) is faster than EfficientNet-B2 (88ms vs. 101ms) on an Intel Xeon CPU E5-2630.
 
\vspace{-4.5mm}

\paragraph{Training Faster RCNN.}
We adopt Faster RCNN~\cite{fasterrcnn} to explore the maximal performance of ReXNets. We plug ReXNets (×0.9) and (×2.2), EfficientNet-B0, and ResNet50 into FPN~\cite{fpn} and train with the image size of $1200\stimes800$ following the original setting~\cite{fasterrcnn,fpn} such as freezing all BNs. Table~\ref{table:coco_frcnn} shows ReXNets' superiority over others; ReXNet (×0.9) and ReXNet (×2.2) improves EfficientNet-B0 and ResNet50 by {\bf +0.5pp} and {\bf +3.9pp} APs, respectively with smaller computational costs; ReXNet (×2.2) achieves {\bf 41.5} AP only with the standard Faster RCNN framework without any bells and whistles. 

\subsection{Fine-grained classifications}
\vspace{-1mm}
We finetune the ImageNet-pretrained models on the datasets Food-101~\cite{food101}, Stanford Cars~\cite{stanford_cars}, FGVC Aircraft~\cite{fgvc_aircraft}, and Oxford Flowers-102~\cite{flower102} to verify the transferability. We compare ReXNet (×1.0) with  ResNet50~\cite{resnet} and EfficientNet-B0~\cite{efficientnet} on each dataset. We exhaustively search the hyper-parameters including learning rate and weight decay for the best results as done in the work~\cite{kornblith2019do_imagenet} for each model. We train all the layers using SGD with the same learning rates without using additional training techniques. 
Training and evaluation are done with $224\stimes224$ image size; we use center-cropped images from the resized images with the shorter side of 256 for evaluation. 
Table~\ref{table:transferlearning} shows ReXNet (×1.0) outperforms EfficientNet-B0 for all the datasets with large margins and mostly surpasses ResNet50 which has {\bf 5×} more parameters. This indicates our backbone can perform as a more generalizable feature extractor than other models even with fewer parameters and may reflect the effectiveness of our channel configuration.

\subsection{COCO Instance Segmentation}
\vspace{-1mm}
We use Mask RCNN~\cite{maskrcnn} to validate the performance of ReXNets on instance segmentation. %
We train the models with the identical setting in the Faster RCNN training in Table~\ref{table:coco_frcnn}. Table~\ref{table:coco_maskrcnn} shows our backbones' efficiency; ReXNet (×0.9) outperforms EfficientNet-B0 by {\bf +0.6pp} mask AP and {\bf +0.5pp} bbox AP with fewer parameters; ReXNet (×2.2) gains {\bf +3.2pp} mask AP and {\bf +3.5pp} bbox AP over ResNet50 with much less computational complexity.

%% file: tables/imagenet_vs_light.tex
\begin{table}[t]
\centering
\fontsize{8.3}{9.3}\selectfont
\tabcolsep=0.25cm
\begin{tabular}{@{}l|cc|cc@{}}
Network  & Top-1 & Top-5 & FLOPs &  Params \\
\midrule
MobileNetV1~\cite{mobilenetv1}          & 70.6\% & 89.5\% & 0.56B & 4.2M \\
MobileNetV2~\cite{mobilenetv2}          & 72.0\% & 91.0\% & 0.30B & 3.5M \\
MobileNetV2 (x1.4)~\cite{mobilenetv2}   & 74.7\% & 92.5\% & 0.59B & 6.9M \\
ShuffleNetV1 (x2)~\cite{shufflenetv1}   & 73.7\% & -    & 0.52B & -    \\
ShuffleNetV2 (x2)~\cite{shufflenetv2}   & 75.4\% & -    & 0.60B & -    \\
\midrule
NASNet-A~\cite{nasneta}                 & 74.0\% & 91.7\% & 0.56B & 5.3M \\
AmoebaNet-A~\cite{amoebanet}            & 75.5\% & 92.0\% & 0.56B & 5.1M \\
PNASNet~\cite{pnasnet}                  & 74.2\% & 91.9\% & 0.59B & 5.1M \\
DARTS~\cite{darts}                      & 73.1\% & 91.0\% & 0.60B & 4.9M \\
FBNet-C~\cite{fbnet}                    & 74.9\% & -    & 0.38B & 5.5M \\
Proxyless-R~\cite{proxylessnas}         & 74.6\% & 92.2\% & 0.32B & 4.1M \\
P-DARTS~\cite{chen2019pdarts}           & 75.6\% & 92.6\% & 0.56B & 4.9M \\
\midrule
MobileNetV3-Large$^\dagger$~\cite{mobilenetv3}    & 75.2\% & -    & 0.22B & 5.4M \\
MnasNet-A3$^\dagger$~\cite{mnasnet}               & 76.7\% & 93.3\% & 0.40B & 5.2M \\
MixNet-M$^{\dagger}$~\cite{tan2019mixconv}          & 77.0\% & 93.3\% & 0.36B & 5.0M \\
EfficientNet-B0$^\dagger$~\cite{efficientnet}     & 76.7\% & - & 0.39B & 5.3M \\
AtomNAS-C+$^{\dagger}$~\cite{mei2020atomnas}        & 77.6\% & 93.6\% & 0.36B & 5.9M \\
EfficientNet-B0$^{\dagger*}$~\cite{efficientnet}     & 77.3\% & - & 0.39B & 5.3M \\
FairNAS-A$^{\dagger*}$~\cite{chu2019fairnas}        & 77.5\% & 93.7\% & 0.39B & 5.3M \\
FairDARTS-C$^{\dagger*}$~\cite{chu2019fairdarts}    & 77.2\% & 93.5\% & 0.39B & 5.3M \\
SE-DARTS+$^{\dagger*}$~\cite{liang2019dartsplus}    & 77.5\% & 93.6\% & 0.59B & 6.1M \\
\midrule
ReXNet (×1.0)$^\dagger$                             & 77.3\% & 93.4\% & 0.40B & 4.8M \\
ReXNet (×1.0)$^{\dagger*}$        & {\bf 77.9}\% & {\bf 93.9}\% & 0.40B & 4.8M \\
\end{tabular}
\vspace{-3mm}
\caption{\small {\bf Comparison of ImageNet performance.} We compare ReXNet (×1.0) with other lightweight models including structured NAS-based models. We report the accuracy at the final epoch of training. $^\dagger$: used SE and SiLU~; $^*$: used additional techniques such as AutoAug, RandAug, or Mixup.}%
\label{table:imagenet_vs_light}
\vspace{-5mm}
\end{table}

%% file: figures/figure__vs_efficientnets.tex
\begin{figure}[t]
    \centering
    \hspace{-4.1mm}
    \begin{subfigure}[t]{0.34\linewidth}
        \includegraphics[page=1, trim = 35mm 6mm 30mm 26mm, clip, width=1.0\linewidth]{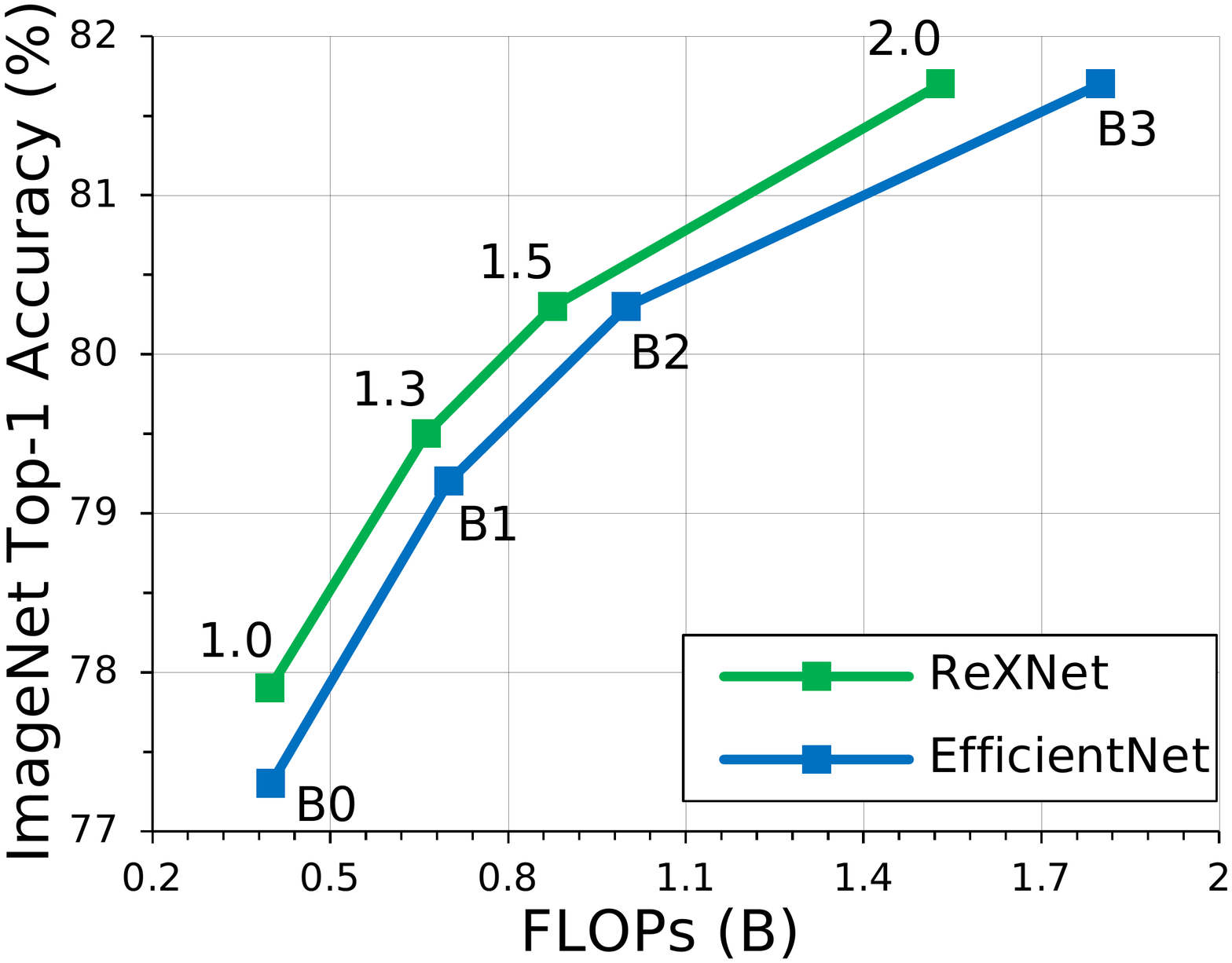} 
    \end{subfigure}
    \begin{subfigure}[t]{0.34\linewidth}
        \includegraphics[page=2, trim = 35mm 6mm 30mm 26mm, clip, width=1.0\linewidth]{figures/sources/figure_set.pdf} 
    \end{subfigure} 
    \begin{subfigure}[t]{0.34\linewidth}
        \includegraphics[page=3, trim = 35mm 6mm 30mm 26mm, clip, width=1.0\linewidth]{figures/sources/figure_set.pdf} 
    \end{subfigure} \\
    \vspace{-3mm}
    \caption{\small {\bf ImageNet accuracy vs. FLOPs and latencies}. We visualize the numbers of ReXNets (×1.0, ×1.3, ×1.5, and ×2.0) and EfficientNets-B0, B1, B2, and B3 in Table~\ref{table:imagenet_vs_efficientnet}. We observe ReXNet outperforms each of the EfficientNet counterparts in practice.}
    \label{fig:vs_efficinetnets}
    \vspace{-2.6mm}
\end{figure} 

%% file: tables/imagenet_vs_eff.tex
\begin{table}[t]
\centering
\fontsize{8.3}{9.3}\selectfont
\tabcolsep=0.15cm
\begin{tabular}{@{}l|cc|rrrr@{}}
Network  & Top-1 & Top-5  & FLOPs  &  Params & CPU & GPU \\
\midrule
ReXNet (×0.9) & {\bf 77.2}\% & {\bf 93.5}\% & 0.35B & 4.1M & 45ms & 20ms\\
\midrule
Eff-B0~\cite{efficientnet}	& 77.3\% & 93.5\% & 0.39B & 5.3M & 47ms & 23ms \\
ReXNet (×1.0) & {\bf 77.9}\% & {\bf 93.9}\% & 0.40B & 4.8M & 47ms & 21ms \\  
\midrule
Eff-B1~\cite{efficientnet}	& 79.2\% & 94.5\% & 0.70B & 7.8M & 70ms & 37ms \\
ReXNet (×1.3) & {\bf 79.5}\% & {\bf 94.7}\% & 0.66B & 7.6M & 55ms & 28ms\\ 
\midrule
Eff-B2~\cite{efficientnet}	& 80.3\% & 95.0\% & 1.0B & 9.2M & 77ms & 48ms \\
ReXNet (×1.5) & {\bf 80.3}\% & {\bf 95.2}\% & 0.9B & 9.7M & 59ms & 31ms\\ 
\midrule
Eff-B3~\cite{efficientnet}	& {\bf 81.7}\% & 95.6\% & 1.8B & 12M & 100ms & 78ms\\
ReXNet (x2.0) & 81.6\% & {\bf 95.7}\% & 1.5B & 16M & 69ms & 40ms \\ 
\end{tabular}
\vspace{-3mm}
\caption{\small {\bf Scalablity of our models}. We adjust ReXNet (×1.0) via width multipliers to compare with EfficientNets~\cite{efficientnet} on the ImageNet dataset. We report the overall performances with CPU and GPU latencies in practice.} %
\label{table:imagenet_vs_efficientnet}
\vspace{-5.5mm}
\end{table}

%% file: tables/coco_overall.tex
\begin{table*}[t]
\centering
\fontsize{8.5}{9.5}\selectfont
\tabcolsep=0.35cm
\begin{tabular}{@{}l|c|ccc|cc@{}}
\multirow{2}{*}{Model} & 
\multirow{2}{*}{Input Size} & 
\multicolumn{3}{c|}{Bbox AP at IOU} &
\multirow{2}{*}{Params} &
\multirow{2}{*}{FLOPs} \\
& & AP & AP$_{\text{50}}$ & AP$_{\text{75}}$ & \\
\midrule
MobileNetV1~\cite{mobilenetv1} + SSDLite                         & 320×320 & 22.2 & -    & -    & 5.1M  & 1.31B \\
MobileNetV2~\cite{mobilenetv2}  + SSDLite                         & 320×320 & 22.1 & -    & -    & 4.3M  & 0.79B \\
MobileNetV3~\cite{mobilenetv3} + SSDLite                          & 320×320 & 22.0 & -    & -    & 5.0M & 0.62B \\
MnasNet-A1~\cite{mnasnet} + SSDLite                           & 320×320 & 23.0 & -    & -    & 4.9M  & 0.84B \\
\midrule
EfficientNet-B0~\cite{efficientnet} + SSDLite$^\dagger$                         & 320×320 & 23.5 & 39.5 & 23.8 & 6.2M  &  0.97B \\
{\bf ReXNet (×0.9) + SSDLite}                           & 320×320 &  24.7 & 41.4 & 24.9 & 5.0M & 0.88B \\
{\bf ReXNet (×1.0) + SSDLite}                            & 320×320 & {\bf 25.3} & {\bf 42.3} & {\bf 25.6} & 5.7M &  1.01B \\
\midrule
EfficientNet-B1~\cite{efficientnet} + SSDLite$^\dagger$                         & 320×320 & 25.8 & 42.5 & 26.2 &  8.7M  &  1.35B \\
EfficientNet-B2~\cite{efficientnet} + SSDLite$^\dagger$                         & 320×320 & 26.6 & 43.6 & 27.2 &  10.0M  &  1.55B \\
{\bf ReXNet (×1.3) + SSDLite}                            & 320×320 & {\bf 27.3} & {\bf 45.2} & {\bf 27.8}   & 8.4M & 1.60B \\
\end{tabular}
\vspace{-3mm}
\caption{\small {\bf COCO object detection results with SSDLite~\cite{mobilenetv2}}. We report box AP scores on \texttt{testdev2017} of our models with SSDLite comparing with lightweight models (FLOPs$\simeq$1.0B). %
$^\dagger$: the model performances are trained by ourselves.}
\label{table:coco_overall}
\vspace{-4mm}
\end{table*}

%% file: tables/coco_frcnn.tex
\begin{table*}[t]
\small
\centering
\fontsize{8.5}{9.5}\selectfont
\tabcolsep=0.4cm
\begin{tabular}{@{}l|c|ccc|cc@{}}
\multirow{2}{*}{Backbone} & 
\multirow{2}{*}{Input Size} & 
\multicolumn{3}{c|}{Bbox AP at IOU} &
\multirow{2}{*}{Params} &
\multirow{2}{*}{FLOPs} \\
& & AP & AP$_{\text{50}}$ & AP$_{\text{75}}$ & \\
\midrule
EfficientNet-B0~\cite{efficientnet} + FPN & 1200×800 & 38.0 & 60.1 & 40.4 & 21.0M  &  123.0B \\
{\bf ReXNet (×0.9)} + FPN    & 1200×800 & {\bf 38.0} & {\bf 60.6} & {\bf 40.8} & 20.1M &  123.0B \\

\midrule
ResNet50~\cite{resnet} + FPN      & 1200×800 & 37.6 & 58.2 & 40.9 & 41.8M &  202.2B \\
{\bf ReXNet (×2.2)} + FPN       & 1200×800 & {\bf 41.5} & {\bf 64.0} & {\bf 44.9} & 33.0M &  153.8B \\
\end{tabular}
\vspace{-3mm}
\caption{\small {\bf COCO object detection results with Faster RCNN~\cite{fasterrcnn} and FPN~\cite{fpn}}. We report box APs on \texttt{val2017}.}%
\label{table:coco_frcnn}
\vspace{-6mm}
\end{table*}

%% file: tables/transferlearning.tex
\begin{table}[t]
\fontsize{8.5}{9.5}\selectfont
\centering
\tabcolsep=0.15cm
\begin{tabular}{@{}l|l|c|cc@{}}
Dataset & Network  & Top-1 & FLOPs & Params \\
\midrule
\multirow{3}{*}{Food-101~\cite{food101}}      & ResNet50~\cite{resnet}              & 87.0\% & 4.1B & 25.6M \\
                               & EfficientNet-B0~\cite{efficientnet} & 87.5\% & 0.4B & 5.3M \\
                               & {\bf ReXNet (×1.0) }                         & {\bf 88.4}\% & 0.4B & 4.8M\\
\midrule
\multirow{3}{*}{Stanford Cars~\cite{stanford_cars}} & ResNet50~\cite{resnet}              & {\bf 92.6}\% & 4.1B & 25.6M \\
                               & EfficientNet-B0~\cite{efficientnet} & 90.7\% & 0.4B & 5.3M \\
                               & {\bf ReXNet (×1.0) }                         & 91.5\% & 0.4B & 4.8M \\
\midrule
\multirow{3}{*}{Aircraft~\cite{fgvc_aircraft}} & ResNet50~\cite{resnet}              & 89.4\% & 4.1B & 25.6M \\
                               & EfficientNet-B0~\cite{efficientnet} & 87.1\% & 0.4B & 5.3M \\
                               & {\bf ReXNet (×1.0) }                         & {\bf 89.5}\% & 0.4B & 4.8M \\
\midrule                               
\multirow{3}{*}{Flowers-102~\cite{flower102}} & ResNet50~\cite{resnet}         & 97.7\% & 4.1B & 25.6M \\
                               & EfficientNet-B0~\cite{efficientnet} & 97.3\% & 0.4B & 5.3M\\
                               & {\bf ReXNet (×1.0) }                         & {\bf 97.8}\% & 0.4B & 4.8M \\                               
\end{tabular}
\vspace{-3mm}
\caption{\small {\bf Transfer learning results on fine-graned datasets}. Our ReXNet (×1.0) is more efficient than ResNet50 and EfficientNet-B0, yet transfers well to the various datasets. } %
\label{table:transferlearning}
\vspace{-5mm}
\end{table}

%% file: tables/coco_maskrcnn.tex
\begin{table*}[t]
\small
\centering
\fontsize{8.5}{9.5}\selectfont
\tabcolsep=0.3cm
\begin{tabular}{@{}l|c|ccc|ccc|cc@{}}
\multirow{2}{*}{Backbone} & 
\multirow{2}{*}{Input Size} & 
\multicolumn{3}{c|}{Mask AP at IOU} &
\multicolumn{3}{c|}{Bbox AP at IOU} &
\multirow{2}{*}{Params} &
\multirow{2}{*}{FLOPs} \\
&  & $\text{AP}$ & $\text{AP}_{\text{50}}$ & $\text{AP}_{\text{75}}$ & $\text{AP}^{\text{bb}}$ & $\text{AP}^{\text{bb}}_{\text{50}}$ & $\text{AP}^{\text{bb}}_{\text{75}}$ & \\
\midrule
EfficientNet-B0~\cite{efficientnet}  + FPN     & 1200×800 & 34.8 & 56.8 & 36.6 & 38.4 & 60.2 & 40.8 & 23.7M  &  123.0B \\
{\bf ReXNet (×0.9)} + FPN        & 1200×800 & {\bf 35.2} & {\bf 57.4} & {\bf 37.1} & {\bf 38.7} & {\bf 60.8} & {\bf 41.6} & 22.8M &  123.0B \\
\midrule
ResNet50~\cite{resnet}  + FPN           &  1200×800 & 34.6 & 55.9 & 36.8 & 38.5 & 59.0 & 41.6 & 44.2M &  206.5B \\
{\bf ReXNet (×2.2)} + FPN        & 1200×800 & {\bf 37.8} & {\bf 61.0} & {\bf 40.2} & {\bf 42.0} & {\bf 64.5} & {\bf 45.6} & 35.6M &  153.8B \\
\end{tabular}
\vspace{-3mm}
\caption{\small {\bf COCO instance segmentation results with Mask RCNN~\cite{maskrcnn} and FPN~\cite{fpn}}. We report box and mask APs on \texttt{val2017}.}
\label{table:coco_maskrcnn}
\vspace{-5mm}
\end{table*}

%% file: 06.Discussion.tex
\vspace{-1mm}
\section{Discussion}
\vspace{-1mm}
\label{section:ablation_and_discussion}
\paragraph{Fixing network depth and searching models.} We further verify the linear channel parameterization by searching for new models under different constraints.  We fix the network depth as 18 and 30 and give the constraints with FLOPs of 30M, 50M, and 70M for 18-depth models; 100M, 150M, and 200M for 30-depth models, respectively. Eventually, Figure~\ref{fig:depthfixed_searched_model_stats} shows all the top 10\% models have identical shapes that are linear functions but have different slopes due to the different FLOPs. This shows that linear channel configurations outperform the conventional configuration for various computational demands.%
\input{figures/figure__depthfixed_channel_search.tex}

\vspace{-4mm}
\paragraph{Rank visualization.} %
We study the model expressiveness by analyzing the rank of trained models. This analysis is to see how the linear parameterization affects the rank. We visualize the rank computed from the trained models in two manners; we show the distribution of accuracy vs. rank (represented by the nuclear norm) from 18-depth models in \S\ref{section:sec4}; we then compare MobileNetV2
with ReXNet (×0.9) by visualizing the cumulative distribution of the singular values that are normalized to [0, 1] computed from the features of the images in the ImageNet validation set. 
As shown in Figure~\ref{fig:rank_expansion}, we observe 1) a higher-accuracy model has a higher rank; 2) ReXNet clearly expands the rank over the baseline.

%% file: figures/figure__depthfixed_channel_search.tex
\begin{figure}[t]
\small
\centering
\hspace{-2mm}
\begin{subfigure}[ht!]{0.5\linewidth}
\includegraphics[page=1, trim = 0mm 0mm 0mm 0mm, clip, width=1.0\linewidth]{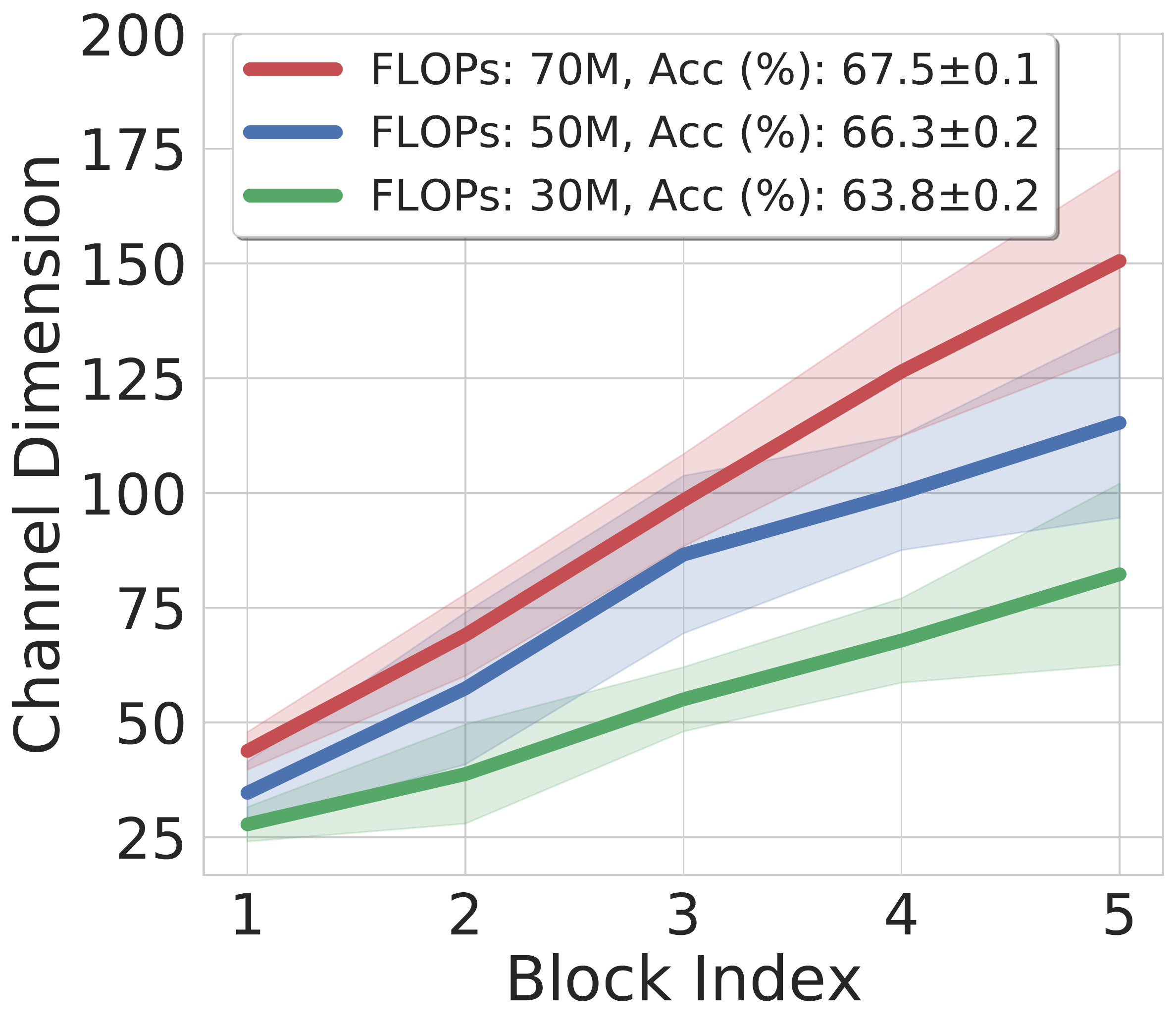} 
\vspace{-6mm}
\caption{\small Top-10\% models (18-depth)}
\end{subfigure}
\begin{subfigure}[ht!]{0.5\linewidth}
\includegraphics[page=1, trim = 0mm 0mm 0mm 0mm, clip, width=1.0\linewidth]{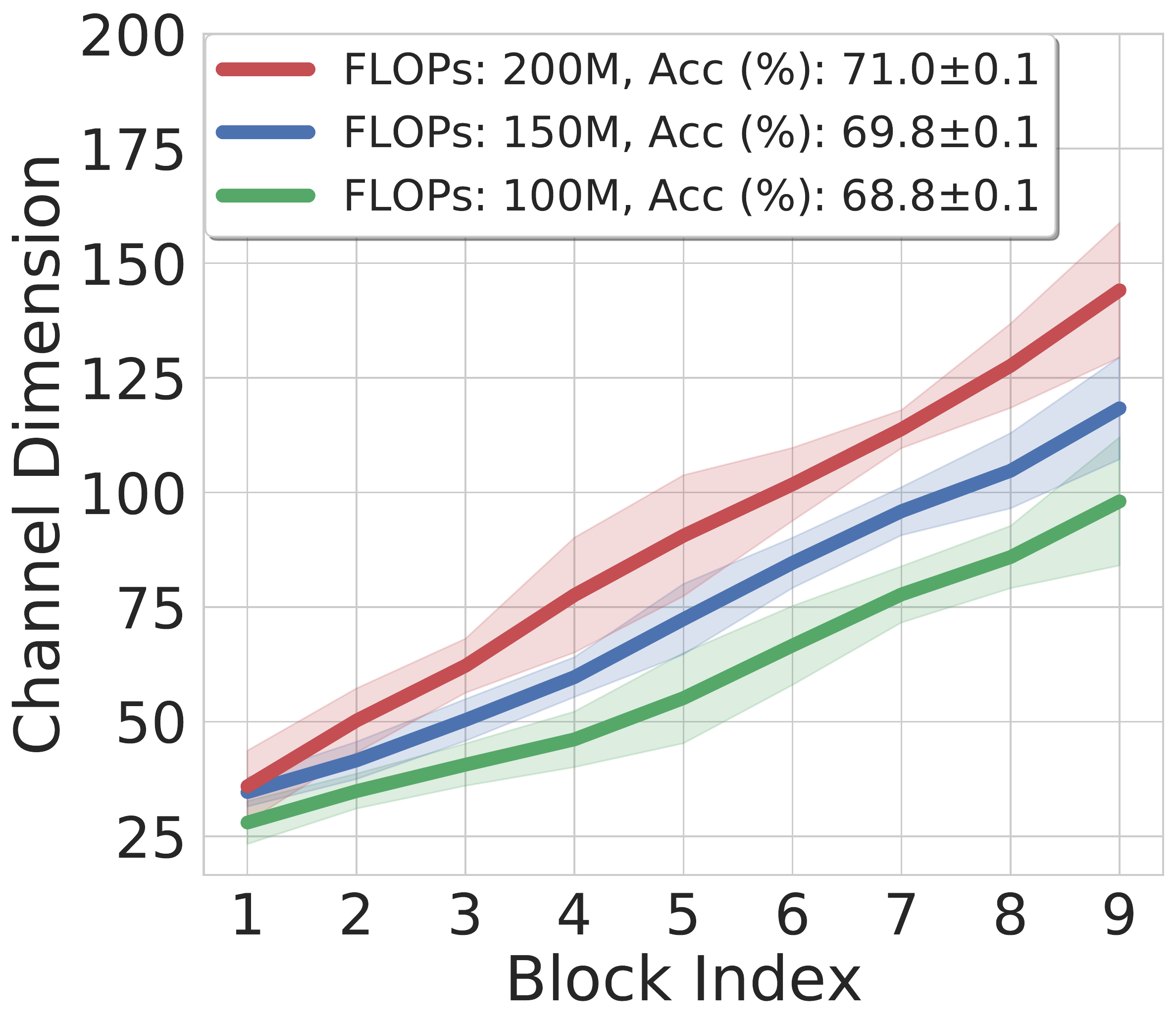} \vspace{-6mm}
\caption{\small Top-10\% models (30-depth)}
\end{subfigure}
\vspace{-3mm}
\caption{\small {\bf Searched configuration under fixed depth}. We similarly plot the searched channel dimensions of top-10\% models from the searched candidates with fixed depths. The search constraints (FLOPs) are {\color{red} Red}: 70M (left) 200M (right); {\color{RoyalBlue} Blue:} 50M (left) and 150M (right); {\color{ForestGreen} Green}: 30M (left) and 100M (right).}%
\label{fig:depthfixed_searched_model_stats}
\vspace{-5mm}
\end{figure}

%% file: 07.Conclusion.tex
\input{figures/figure__model_expressiveness}
\vspace{-1mm}
\section{Conclusion}
\vspace{-1mm}
\label{section:conclusion}
In this work, we have studied a new approach to designing lightweight models. We have conjectured that the conventional stage-wise channel configurations which have widely used in NAS methods after proposed by MobileNetV2 have limited the model accuracy; if we find a more effective channel configuration, the models would have accuracy gains. We first studied an appropriate way of designing a single building block and a layer targeted to lightweight models; we then proposed a search method for a channel configuration via piece-wise linear functions of block index. The search space contains the conventions, and we found an effective channel configuration which can be parameterized by a linear function of the block index. Based on the parameterization, we have achieved a new model which could outperform the recent lightweight models including NAS-based models on ImageNet dataset. Our ReXNets further showed remarkable finetuning performances on COCO object detection, instance segmentation, and fine-grained classifications. Consequently, we believe our work highlighted a new perspective of architecture search that helps to search with.

%% file: figures/figure__model_expressiveness.tex
\begin{figure}[t]
\small
\centering
\hspace{-2.5mm}
\begin{subfigure}[ht!]{0.5\linewidth}
\includegraphics[page=1, trim = 0mm 0mm 0mm 0mm, clip, width=1.0\linewidth]{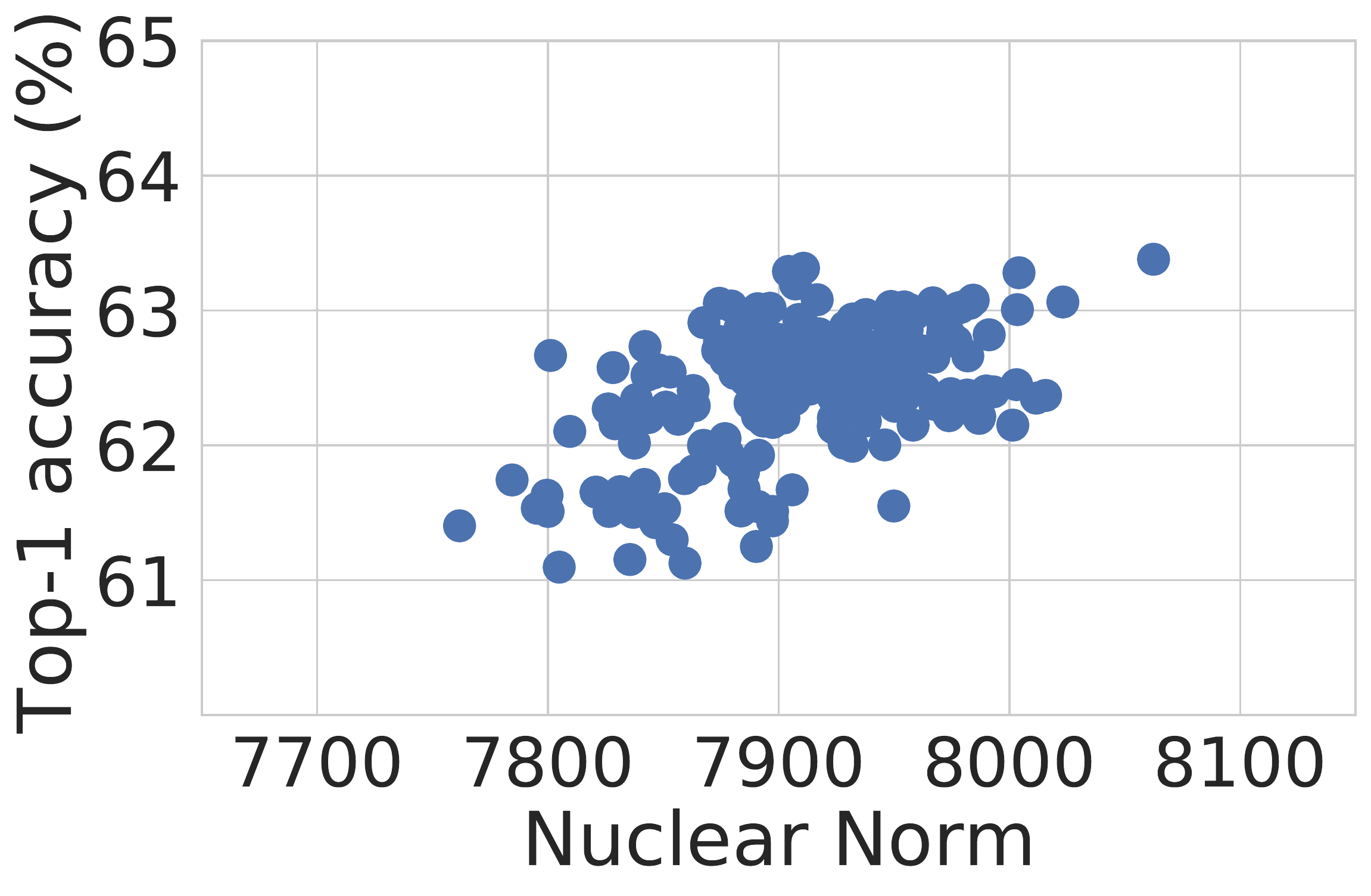} 
\end{subfigure}
\begin{subfigure}[ht!]{0.5\linewidth}
\includegraphics[page=1, trim = 0mm 0mm 0mm 0mm, clip, width=1.0\linewidth]{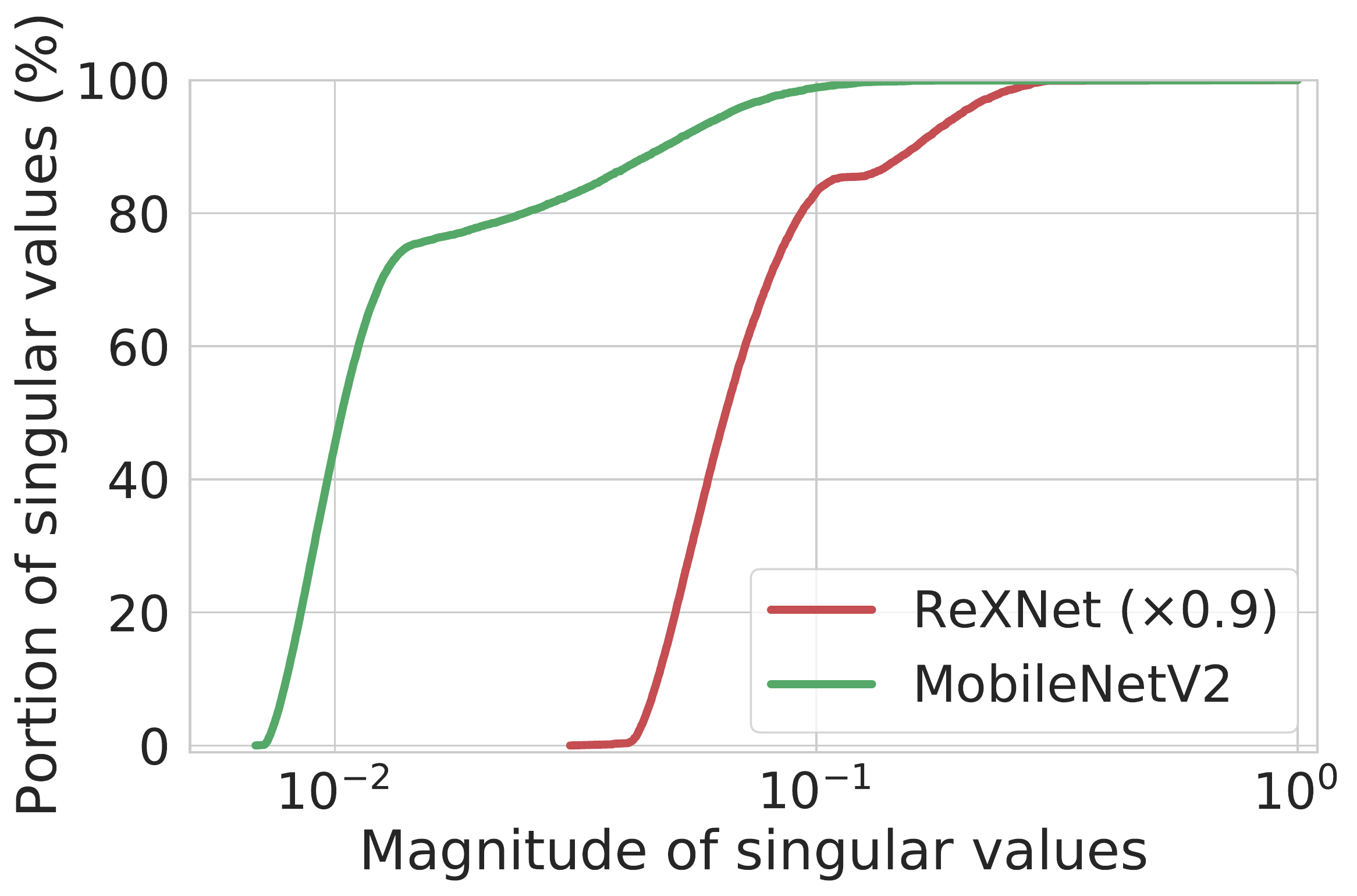} 
\end{subfigure}
\vspace{-3mm}
\caption{\small {\bf Visualization of rank}. The models with higher accuracy have higher nuclear norms (left); our ReXNet has larger singular values than MobileNetV2 (right).}
\label{fig:rank_expansion}
\vspace{-6mm}
\end{figure}

%% file: appendix.tex
\maketitle

\appendix
\renewcommand{\thefigure}{A\arabic{figure}}
\setcounter{figure}{0}
\renewcommand{\thetable}{A\arabic{table}}
\setcounter{table}{0}

\input{supplementary_source/s2.search_result}

\input{supplementary_source/s3.model_specification}

\input{supplementary_source/s4.further_results}

%% file: supplementary_source/s2.search_result.tex
\section{Additional Search Results}
\label{app:cifar10_search}
We provide the additional results of searching an effective channel dimension configuration by the proposed search method described in \S\ref{section:sec4} to show the consistent trend of the searched channel configurations over different datasets. We perform searches on the CIFAR-10 dataset~\cite{cifar} with the identical search constraints presented in Table~\ref{table:channel_search_result}. %

Following the previous analysis, we collect top-10\%, middle-10\% (i.e., the models between the top 50\% and 60\%), and bottom-10\% models in terms of the model accuracy from 200 searched models to show the channel configurations with the performance statistics after each search. Figure~\ref{supp_fig:cifar10_searched_model_stats} shows that the searched models as colored with {\color{red} red}, which look linear functions, have higher accuracy while maintaining the similar computational costs. These similar trends are regularly observed while searching under the various constraints, and we can parameterize the models with a linear function by the block index again. The models in {\color{ForestGreen} green} have highly reduced the input-side channels and many output-side weight parameters resulting in the loss of accuracy. In addition, {\color{RoyalBlue} blue} represents the models at the middle-10\% accuracy, which looks similar to the conventional channel configurations such as MobileNetV2's~\cite{mobilenetv2}. 

All the searched channel configurations which can be approximated to linear parameterizations by the block index have higher accuracy ({\color{red} red}) then {\color{RoyalBlue} blue} ones, which are show the identical trends to Figure~\ref{fig:searched_model_stats}. %
It is worth noting that all the {\color{red} red} lines in Figure~\ref{supp_fig:cifar10_searched_model_stats} have higher slopes compared to those in Figure~2 in the main paper. %
This is because CIFAR-100 models have more parameters at the final classifier due to a larger number of classes, so the early layers employ fewer parameters than those of the models trained on the CIFAR-10 dataset.
\input{supplementary_source/figures/figure__cifar10_searched_model}

\input{supplementary_source/figures/figure__model_blocks}

%% file: supplementary_source/figures/figure__cifar10_searched_model.tex
\begin{figure}[t]
\small
\centering
\hspace{-2mm}
\begin{subfigure}[ht!]{0.48\linewidth}
\includegraphics[page=1, trim = 0mm 0mm 0mm 0mm, clip, width=1.0\linewidth]{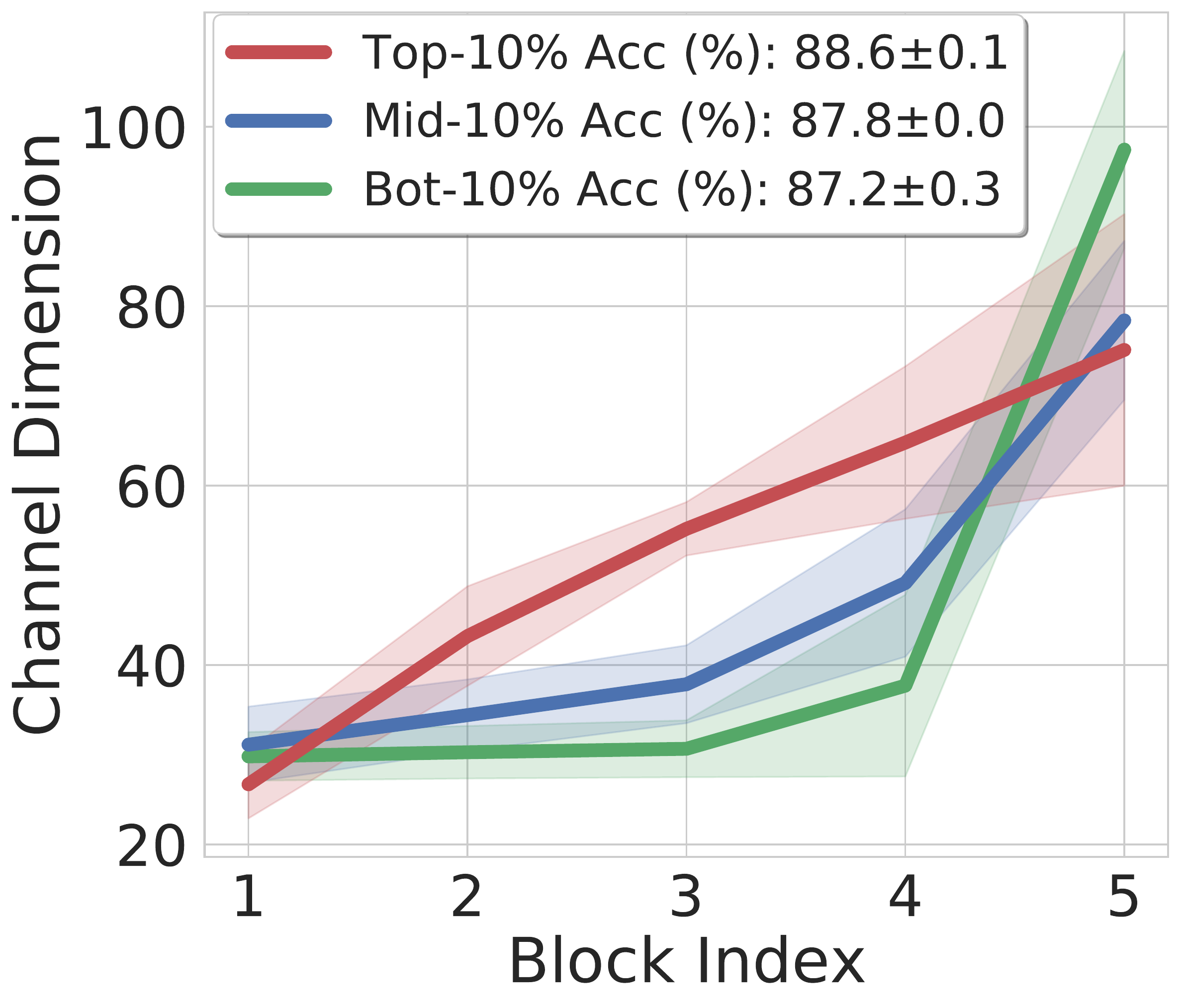} 
\vspace{-6mm}
\caption{\fontsize{8.0}{9}\selectfont
 Depth: 18 (\# inverted bot.: 5),\\ \# Params$\simeq$0.2M, FLOPs$\simeq$30M}
\end{subfigure}
\quad 
\begin{subfigure}[ht!]{0.48\linewidth}
\includegraphics[page=1, trim = 0mm 0mm 0mm 0mm, clip, width=1.0\linewidth]{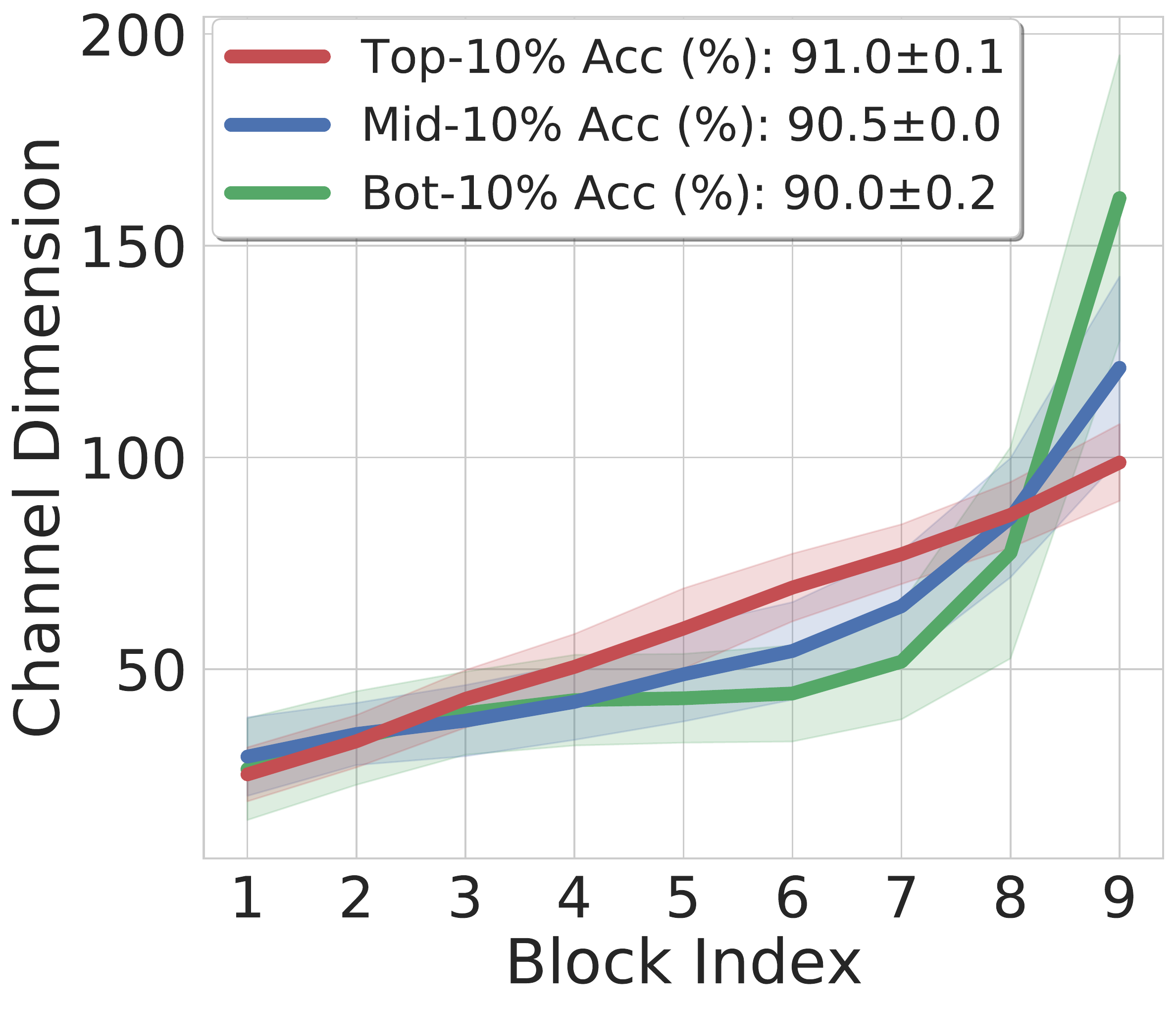}
\vspace{-6mm}
\caption{\fontsize{8.0}{9}\selectfont 
 Depth: 30 (\# inverted bot.: 9),\\ \# Params$\simeq$0.5M, FLOPs$\simeq$100M}%
 \label{subfig:depth11}
\end{subfigure} 
\begin{subfigure}[ht!]{0.47\linewidth}
\includegraphics[page=1, trim = 0mm 0mm 0mm 0mm, clip, width=1.0\linewidth]{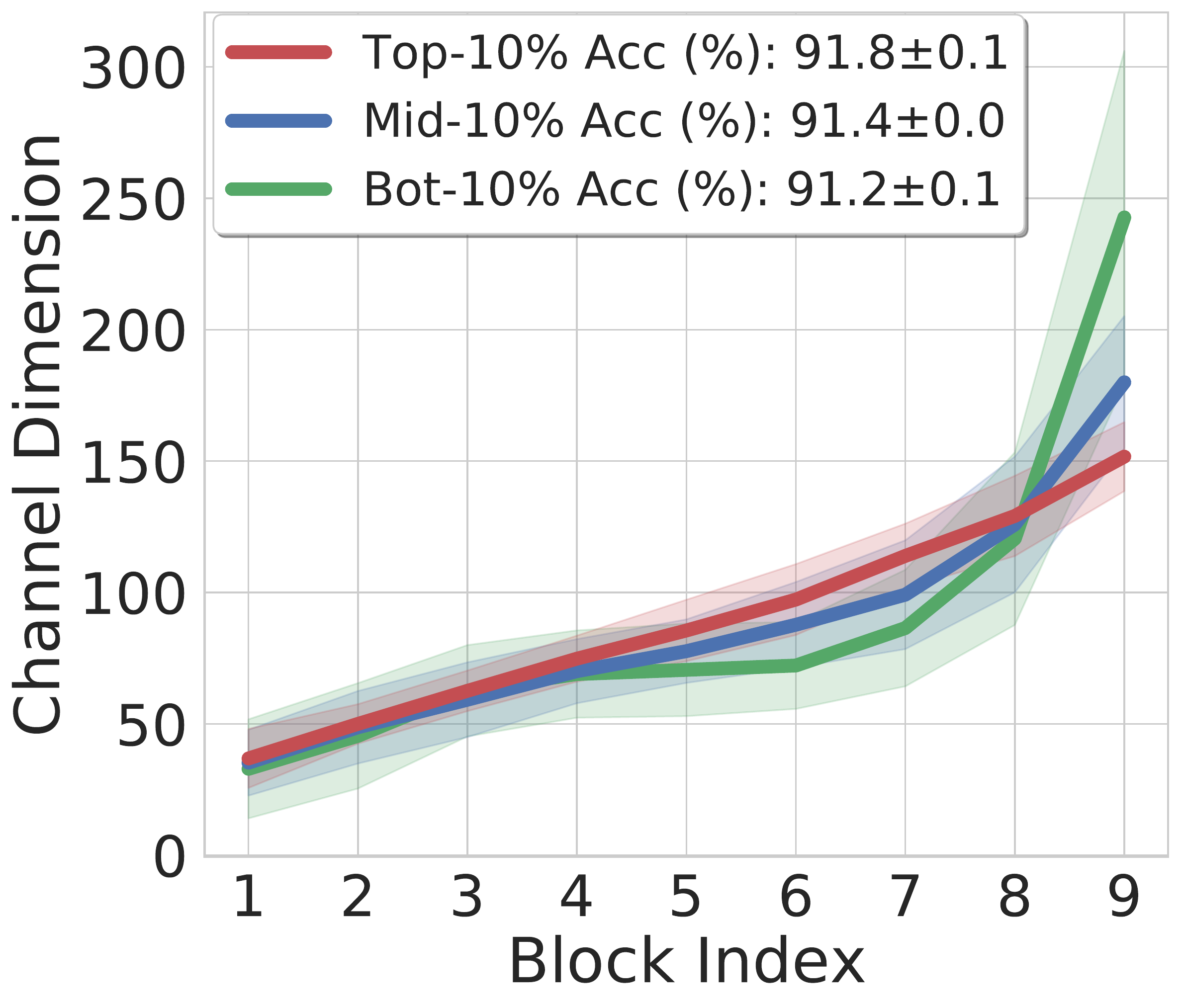} 
\vspace{-6mm}
\caption{\fontsize{8.0}{9}\selectfont
 Depth: 30 (\# inverted bot.: 9),\\ \# Params$\simeq$1.0M, FLOPs$\simeq$200M}
\end{subfigure}
\quad 
\begin{subfigure}[ht!]{0.47\linewidth}
\includegraphics[page=1, trim = 0mm 0mm 0mm 0mm, clip, width=1.0\linewidth]{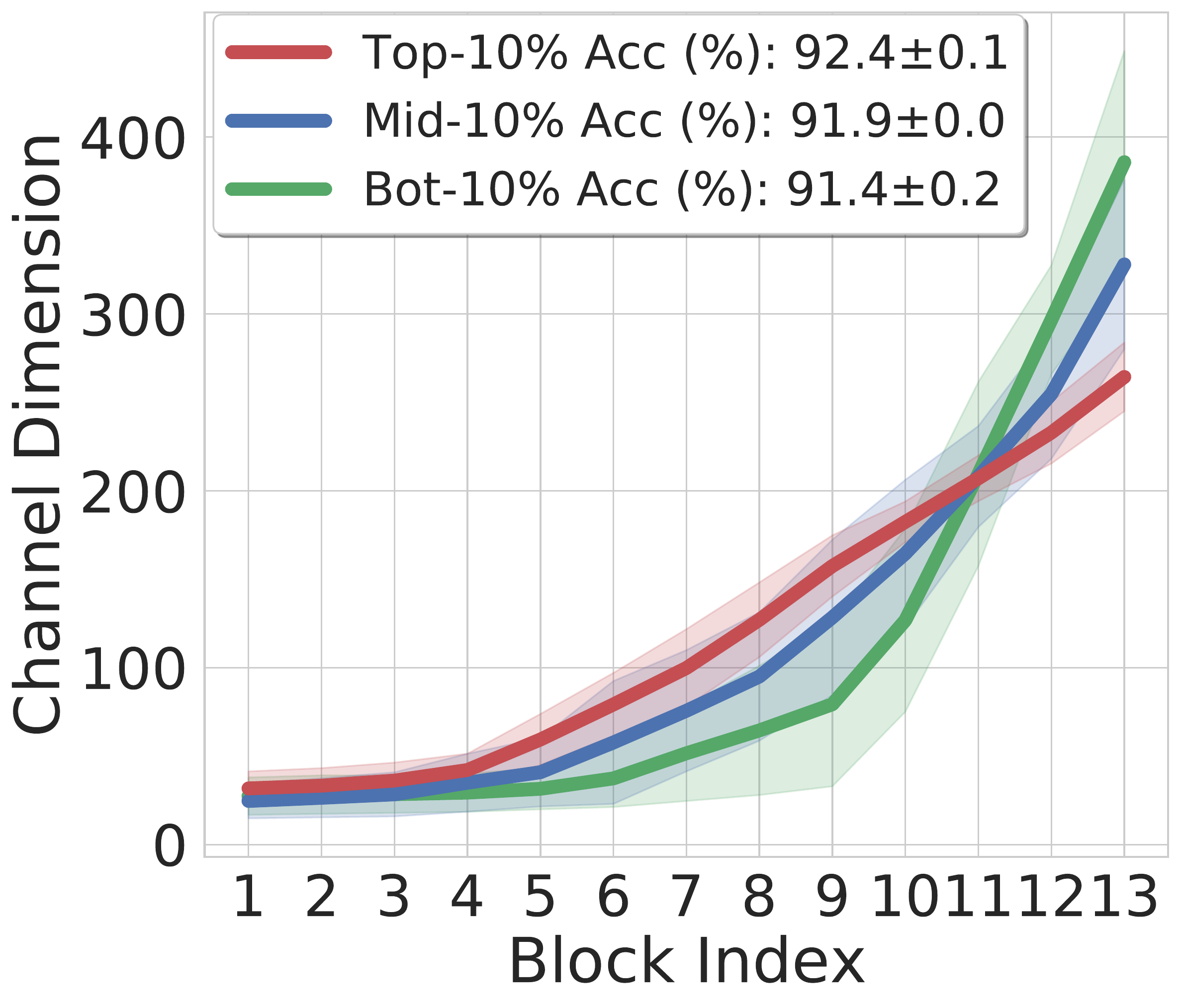}
\vspace{-6mm}
\caption{\fontsize{8.0}{9}\selectfont
 Depth: 42 (\# inverted bot.: 13),\\ \# Params$\simeq$3.0M, FLOPs$\simeq$350M}
\end{subfigure}
\vspace{-2mm}
\caption{\small {\bf Visualization of searched models' channel dimension on CIFAR-10}.   {\color{red} Red}: top 10\%-accuracy models; {\color{RoyalBlue} Blue:} middle 10\% models; {\color{ForestGreen} Green}: bottom 10\% models; we plot the averaged channel configurations with the 1-sigma range and report the averaged top-1 accuracy over each searched candidate.}%
\label{supp_fig:cifar10_searched_model_stats}
\vspace{-3mm}
\end{figure}

%% file: supplementary_source/figures/figure__model_blocks.tex
\begin{figure*}[t]
\small
\centering
\begin{subfigure}[ht!]{1.0\linewidth}
\includegraphics[page=1, trim = 14mm 145mm 90mm 37mm, clip, width=1.0\linewidth]{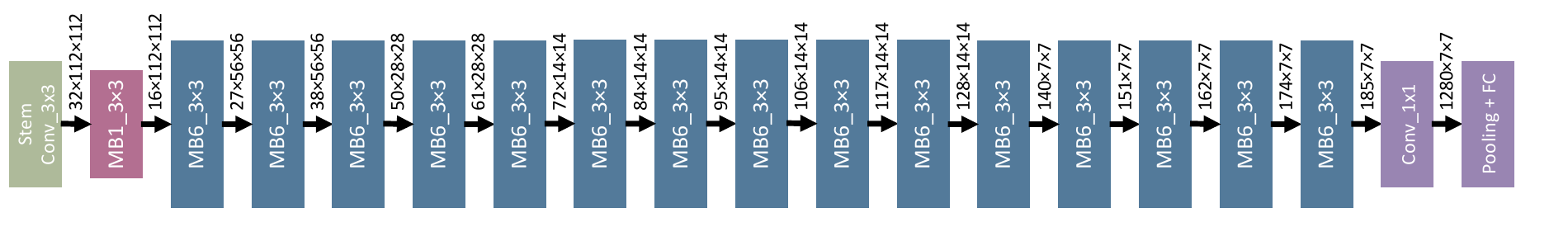}
\vspace{-6mm}
\caption{\small ReXNet (×1.0)}
\label{subfig:our_block}
\end{subfigure} 
\begin{subfigure}[ht!]{1.0\linewidth}
\includegraphics[page=2, trim = 14mm 145mm 90mm 37mm, clip, width=1.0\linewidth]{figures/sources/model_blocks.pdf}
\vspace{-6mm}
\caption{\small ReXNet (plain)}
\label{subfig:plain_block}
\end{subfigure}
\vspace{-2mm}
\caption{\small {\bf Architectures of ReXNet (×1.0) and ReXNet (plain)}. MB1 and MB6 refer to MobileNetV2~\cite{mobilenetv2}'s inverted bottlenecks with the expansion ratio of 1 and 6, respectively. Each model has almost similar architectural elements compared to the original ones.}
\label{supp_fig:model_block}
\vspace{-3mm}
\end{figure*}

%% file: supplementary_source/s3.model_specification.tex
\section{Network Upgrade (cont'd)}
\label{app:models}
In this section, we give further information of ReXNets and introduce our new model rebuilt upon MobileNetV1~\cite{mobilenetv1} called ReXNet (plain) which does not use skip connections~\cite{resnet, mobilenetv2} at each building block. 

\subsection{ReXNet (cont'd)}
We have described our upgraded model based on MobileNetV2~\cite{mobilenetv2}, which follows the searched linear parameterization on channel dimensions with some minor modifications in \S4.4. %
Here, we illustrate the network architecture of our ReXNet (×1.0) in Figure~\ref{subfig:our_block}. We observe ReXNet (×1.0) has the identical block configuration to that of MobileNetV2 where a single-type building block MB6\_3x3, which is the original inverted bottleneck~\cite{mobilenetv2} with the 3×3 depthwise convolution and the expansion ratio 6 is used as the basic building block except for the first inverted bottleneck. Every inverted bottleneck block that expands the channel dimensions (except for the downsampling blocks) has a skip connection where the expanded channel dimensions are padded with zeros. 

\subsection{ReXNet (plain)}
We now present a new model redesigned based on MobileNetV1~\cite{mobilenetv1}. We choose MobileNetV1 as another baseline because we intend to show a network architecture without skip connections (so-called a plain network) is able to be redesigned by following the proposed linear parameterization to show performance improvement. We do not change the depth of MobileNetV1. We use the identical configuration at the stem (i.e., 3×3 convolution with BN and ReLU) and the same large expansion layer at the penultimate layer. 

We reassign the output channel dimensions of each 1×1 convolution as we did for ReXNet in \S\ref{subsection:network_upgrade}. %
Following the investigation of single-layer design, we only replace the ReLUs with SiLU~\cite{gelu,swish} after the expansion layers such as all the $1\stimes1$ convolutions. We leave the ReLUs right after each depthwise convolution where the channel dimension ratio is 1. All the other channel dimensions including the stem and the penultimate layer are not changed. Since the network is a plain network, we do not adopt SE~\cite{SENet}. Our ReXNet (plain) is illustrated in Figure~\ref{subfig:plain_block}. We provide the ImageNet performance of ReXNet (plain) in Table~\ref{supp_table:imagenet_scaled}. We train the model by following the identical training setup in \S5.1. %
As shown in Table~\ref{supp_table:imagenet_scaled}, ReXNet (plain) does not achieve the best accuracy, but it is extremely faster than ReXNets on both CPU and GPU even with larger FLOPs. 

\input{supplementary_source/tables/imagenet_scalablity.tex}

\subsection{Overall models}
\label{app_sub:model_models}
In addition to the models introduced in Table~\ref{table:imagenet_vs_efficientnet}, %
we provide additional models adjusted by different width multipliers. Table~\ref{supp_table:imagenet_scaled} shows the ReXNets (×1.1, ×1.2, ×1.4, ×2.2, and ×3.0) and our new model ReXNet (plain) with the corresponding performances.

\subsection{ReXNet-lite}
We additionally provide faster models based on ReXNet. We make slight changes upon ReXNet: 1) removing SE~\cite{SENet} from each inverted bottleneck; 2) replacing all SiLUs~\cite{gelu, swish} with ReLU6s. We further incorporate a fully connected layer before the classifier. We compare our models with EfficientNet-lites\footnote{https://github.com/tensorflow/tpu/tree/master/models/official/efficientnet/lite} which are lighter version of EffciientNets~\cite{efficientnet} in Table~\ref{supp_table:imagenet_vs_efficientnet_lite}. We measure CPU and GPU inference speeds by averaging the latencies over 1,000 runs with the batch size 1 on an Intel Xeon CPU E5-2630 and the batch size 64 on a V100 GPU, respectively. As shown in the table, ReXNet-lite clearly show better accuracy and faster speed than each of the counterparts in practice.

\input{supplementary_source/tables/imagenet_comparison_lite}

\input{supplementary_source/tables/imagenet_nonlinearity.tex}

%% file: supplementary_source/tables/imagenet_scalablity.tex
\begin{table}[t]
\centering
\fontsize{8.5}{9.5}\selectfont
\tabcolsep=0.15cm
\begin{tabular}{@{}l|cc|rrrr@{}}
Network  & Top-1 & Top-5  & FLOPs  &  Params. & CPU & GPU\\
\midrule
ReXNet (plain) & 74.8\% & 91.9\% & 0.56B & 3.4M  & 22ms & 10ms \\
\midrule
ReXNet (×0.9) & 77.2\% & 93.5\% & 0.35B & 4.1M & 46ms & 20ms\\
ReXNet (×1.0) & 77.9\% & 93.9\% & 0.40B & 4.8M & 47ms & 21ms\\  
ReXNet (×1.1) & 78.6\% & 94.1\% & 0.48B & 5.6M & 51ms & 24ms\\
ReXNet (×1.2) & 79.0\% & 94.3\% & 0.57B & 6.6M & 53ms & 26ms\\
ReXNet (×1.3) & 79.5\% & 94.7\% & 0.66B & 7.6M & 55ms & 28ms\\ 
ReXNet (×1.4) & 79.8\% & 94.9\% & 0.76B & 8.6M & 57ms & 30ms\\ 
ReXNet (×1.5) & 80.3\% & 95.2\% & 0.86B & 9.7M & 59ms & 31ms\\ 
\midrule
ReXNet (×2.0) & 81.6\% & 95.7\% & 1.5B & 16M & 69ms & 40ms\\ 
ReXNet (×2.2) & 81.7\% & 95.8\% & 1.8B & 19M & 73ms & 46ms\\ 
ReXNet (×3.0) & 82.8\% & 96.3\% & 3.4B & 34M & 96ms & 61ms\\ 
\end{tabular}
\vspace{-2mm}
\caption{\small {\bf Performance of ReXNets}. We report the ImageNet~\cite{imagenet} performances of ReXNets. In addition to Table~5 in the main paper, %
we provide more models including ReXNet (plain) and ReXNets (×1.1, ×1.2, ×1.4, ×2.2, ×3.0). All the models are trained and evaluated with the resolution $224\stimes224$.}
\label{supp_table:imagenet_scaled}
\vspace{-3mm}
\end{table}

%% file: supplementary_source/tables/imagenet_comparison_lite.tex
\begin{table}[t]
\centering
\fontsize{8.3}{9.3}\selectfont
\tabcolsep=0.1cm
\begin{tabular}{@{}l|c|rrccc@{}}
Network  & Top-1 & FLOPs  &  Params & CPU & GPU$_{\text{FP32}}$ & GPU$_{\text{FP16}}$\\
\midrule
Eff-lite0     	& 75.1\%  & 0.4B & 4.7M & 30ms & 21ms & 15ms \\
ReX-lite (×1.0) & {\bf 76.2}\% & 0.4B & 4.7M & 31ms & 22ms & 14ms \\  
\midrule
Eff-lite1	    & 76.7\% & 0.6B & 5.4M & 44ms & 32ms & 22ms \\
ReX-lite (×1.3) & {\bf 77.8}\% & 0.6B & 6.8M & 36ms & 28ms & 18ms\\ 
\midrule
Eff-lite2	    & 77.6\% & 0.9B & 6.1M & 48ms & 41ms & 29ms\\
ReX-lite (×1.5) & {\bf 78.6}\% & 0.8B & 8.3M & 39ms & 33ms & 20ms\\ 
\midrule
Eff-lite3	    & 79.8\% & 1.4B & 8.2M & 60ms & 58ms & 42ms\\
ReX-lite (x2.0) & {\bf 80.2}\% & 1.5B & 13.0M & 49ms & 46ms & 26ms\\ 
\end{tabular}
\vspace{-2mm}
\caption{\small {\bf Performance of ReXNet-lites}. We further report the ImageNet performances of ReXNet-lites. We compare with EfficientNet-lites and report the overall performances with CPU and GPU (FP32 and FP16) latencies in practice. } %
\label{supp_table:imagenet_vs_efficientnet_lite}
\vspace{-1mm}
\end{table}

%% file: supplementary_source/tables/imagenet_nonlinearity.tex
\begin{table}[t]
\fontsize{8.5}{9.5}\selectfont
\centering
\tabcolsep=0.15cm
\begin{tabular}{@{}l|cc|cc@{}}
Nonlinearity  & Top-1 (\%) & Top-5 (\%)  &  FLOPs & Params.\\
\midrule
ReLU6~\cite{mobilenetv2} & 77.3 & 93.5 & 0.40B & 4.8M \\
Leaky ReLU~\cite{leakyrelu} & 77.4 & 93.6 & 0.40B & 4.8M \\
Softplus~\cite{softplus} & 77.6  & 93.8 & 0.40B & 4.8M \\
ELU~\cite{elu} & 77.6  & 93.7 & 0.40B & 4.8M \\
SiLU~\cite{gelu,swish} & {\bf 77.9} & {\bf 93.9} & 0.40B & 4.8M \\ 
\end{tabular}
\vspace{-2mm}
\caption{\small {\bf Nonlinear functions and ImageNet accuracy}.}
\label{supp_table:imagenet_nonlinearity}
\vspace{-3mm}
\end{table}

%% file: supplementary_source/s4.further_results.tex
\input{supplementary_source/tables/coco_vs_noisystudent.tex}

\input{supplementary_source/tables/coco_from_scratch_training.tex}

\section{Further Empirical Studies}
\label{app:additional_exp}

\subsection{Impact of nonlinear functions.} We have studied how nonlinearity can affect rank in the investigation in S\ref{section:sec3}. %
We further study the actual impact of them by training the models on ImageNet. We train ReXNet (×1.0) with ELU, SoftPlus, LeakyReLU, ReLU6, and SiLU (Swish-1) with the identical training setup. As shown in Table~\ref{supp_table:imagenet_nonlinearity}, we obtain the results of top-1 accuracy in the order of SiLU (77.9\%), ELU (77.6\%), SoftPlus (77.6\%), Leaky ReLU (77.4\%), and ReLU6 (77.3\%), and the trend is similar to the result in the empirical study in \S\ref{subsection:sec_empirical_study}. %
The result indicates the quality of different nonlinearities that relates to model expressiveness; SiLU shows the best performance. This may provide a backup for why the recent lightweight models use SiLU (Swish-1) as the nonlinearity.

\subsection{COCO object detection (cont'd)}
\label{app:coco_detection}
We further provide more comparisons of ReXNets with EfficientNets~\cite{efficientnet} in SSDLite~\cite{mobilenetv2} on object detection. We first replace the EfficientNet backbone used in \S\ref{subsection:coco_detection} %
with a stronger EfficientNet~\cite{xie2020self}. We then compare them trained from scratch on the COCO dataset~\cite{coco2017}.

\vspace{-3mm}
\paragraph{Comparison with NoisyStudent EfficientNets.}
We now compare ReXNets with stronger EfficientNets~\cite{xie2020self}, where the backbones are trained by a self-training method with extra large-scale data and RandAug~\cite{cubuk2019randaugment}. Our goal is to show ReXNet's architectural capability over the EfficientNets without using the extra data when applying the backbones to a downstream task. We borrow the AP scores on \texttt{val 2017} from the ReXNet+SSDLite models in Table~\ref{table:coco_overall} %
and train EfficientNet-B0, B1, and B2 in SSDLite using the pretrained NoisyStudent+RA EfficientNet models which are publicly released\footnote{https://github.com/tensorflow/tpu/tree/master/models/official/efficientnet}. All the AP scores are evaluated by each checkpoint cached at the last iteration. Table~\ref{supp_table:coco_fromscratch} shows that ReXNets outperform the counterparts with the comparable computational costs. This indicates ReXNets pretrained on ImageNet are still promising.

\vspace{-2mm}
\paragraph{Comparison of the models trained from scratch.}
We aim to verify the model expressiveness itself without using ImageNet-pretrained backbones. This is because one may wonder using a pretrained models trained with different training setups such as optimizer or regularizers affect the performance of downstream tasks. As shown in the work~\cite{he2019rethinking}, the COCO dataset~\cite{coco2017} is able to be trained from scratch, we verify the model's pure expressiveness by training the models from scratch.

We individually train ReXNets (×0.9, ×1.0, and  ×1.3) and EfficientNet-B0, B1, and B2 in SSDLite without using ImageNet pretrained backbones. All AP scores are evaluated by each checkpoint cached at the last iteration again. Table~\ref{supp_table:coco_fromscratch} shows that ReXNets outperform the counterparts by {\bf +1.4pp}, {\bf +0.3pp}, and {\bf +1.2pp} in AP score. With similar computational demands, ReXNets beat the EfficientNet counterparts by large margins, and surprisingly, the training from scratch makes ReXNet (×1.0) outperforms EfficienetNet-B1 by {\bf +0.3pp} with much less computational costs. This indicates that our models are more powerful in terms of expressiveness even without the aids of the supervision of ImageNet pretrained models.

%% file: supplementary_source/tables/coco_vs_noisystudent.tex
\begin{table*}[t!]
\centering
\fontsize{8.5}{9.5}\selectfont
\tabcolsep=0.35cm
\begin{tabular}{@{}l|c|ccc|cc@{}}
\multirow{2}{*}{Model} & 
\multirow{2}{*}{Input Size} & 
\multicolumn{3}{c|}{Bbox AP at IOU} &
\multirow{2}{*}{Params} &
\multirow{2}{*}{FLOPs} \\
& & AP & AP$_{\text{50}}$ & AP$_{\text{75}}$ & \\
\midrule
EfficientNet-B0~\cite{efficientnet} + SSDLite            & 320×320 & 23.6 & 39.4 & 23.3 & 6.2M  &  0.97B \\
{\bf ReXNet (×0.9) + SSDLite}                            & 320×320 & 24.6 & 41.2 & 24.6 & 5.0M & 0.88B \\
\midrule
EfficientNet-B1~\cite{efficientnet} + SSDLite            & 320×320 & 25.6 & 42.2 & 25.8 &  8.7M  &  1.35B \\
{\bf ReXNet (×1.0) + SSDLite}                            & 320×320 & 25.2 & 41.9 & 25.3 & 5.7M &  1.01B \\
\midrule
EfficientNet-B2~\cite{efficientnet} + SSDLite            & 320×320 & 26.4 & 43.4 & 26.6 &  10.0M  &  1.55B \\
{\bf ReXNet (×1.3) + SSDLite}                            & 320×320 & 27.1 & 44.7 & 27.4 & 8.4M & 1.60B \\
\end{tabular}
\vspace{-2mm}
\caption{\small {\bf ReXNets vs. Noisy Student EfficientNets on COCO object detection}. We compare our ReXNets trained solely on ImageNet with stronger EfficientNets trained by Noisy Student training method~\cite{xie2020self} with RandAug~\cite{cubuk2019randaugment}. Note that ReXNets here are equivalent to the models in \S5.2. 
We report box APs on \texttt{val2017}. }
\label{table:coco_noisystudent}
\vspace{-3mm}
\end{table*}

%% file: supplementary_source/tables/coco_from_scratch_training.tex
\begin{table*}[t!]
\small
\centering
\fontsize{8.5}{9.5}\selectfont
\tabcolsep=0.35cm
\begin{tabular}{@{}l|c|ccc|cc@{}}
\multirow{2}{*}{Model} & 
\multirow{2}{*}{Input Size} & 
\multicolumn{3}{c|}{Avg. Precision at IOU} &
\multirow{2}{*}{Params.} &
\multirow{2}{*}{FLOPs} \\
& & AP & AP$_{\text{50}}$ & AP$_{\text{75}}$ & \\
\midrule
EfficienetNet-B0~\cite{efficientnet} + SSDLite          & 320x320 & 23.9 & 39.6 & 24.1& 6.2M  &  0.97B \\
{\bf ReXNet (×0.9) + SSDLite}                           & 320x320 & 25.3 & 41.4 & 25.9 & 5.0M & 0.88B \\
\midrule
EfficienetNet-B1~\cite{efficientnet} + SSDLite          & 320x320 & 25.6 & 41.9 & 26.0 & 8.7M  &  1.35B \\
{\bf ReXNet (×1.0) + SSDLite}                           & 320x320 & 25.9 & 42.6 & 26.3 & 5.7M &  1.01B \\
\midrule
EfficienetNet-B2~\cite{efficientnet} + SSDLite          & 320x320 & 26.5 & 43.3 & 26.7 &  10.0M  &  1.55B \\
{\bf ReXNet (×1.3) + SSDLite}                           & 320x320 & 27.7& 45.1 & 28.0 & 8.4M &  1.60B \\
\end{tabular}
\vspace{-2mm}
\caption{\small {\bf ReXNets vs. EfficientNets on COCO object detection.}. Note that all the models are trained from scratch with the identical training setup except for the doubled training iterations. We report box APs on \texttt{val 2017}.}
\label{supp_table:coco_fromscratch}
\vspace{-3mm}
\end{table*}